\pdfoutput=1
\documentclass[10pt,journal,compsoc]{IEEEtran}

\usepackage{import}
\usepackage{microtype}
\usepackage{graphicx} 
\usepackage{booktabs} 

\usepackage{xspace} 

\usepackage{tikz}
\usetikzlibrary{positioning} 
\usepackage{pgfplots}

\usepackage{amsmath, amssymb, amsfonts} 

\usepackage[thmmarks, amsmath, thref]{ntheorem} 
\usepackage{bbm}
\usepackage{bm}

\usepackage[normalem]{ulem} 
\usepackage{pifont}

\usepackage{newtxtext,newtxmath}

\usepackage{caption}
\usepackage{subcaption}

\usepackage{floatrow}

\usepackage[linesnumbered, algoruled, noend, noline]{algorithm2e}

\usepackage{diagbox}
\usepackage{hyperref}

\hypersetup{
    colorlinks=true, 
    linkcolor=blue,  
    citecolor=blue,  
    filecolor=magenta, 
    urlcolor=blue    
}

\usepackage{enumitem}  
 
\subimport{./}{macros}
\newtheorem{Definition}{Definition}

\usepackage[numbers]{natbib}

\pgfplotsset{compat=1.18}
\begin{document}
\title{Diffeomorphic Temporal Alignment Nets for Time-series Joint Alignment and Averaging}

\author{Ron Shapira Weber
        and~Oren Freifeld}

\markboth{Diffeomorphic Temporal Alignment Networks}%
{}

\IEEEtitleabstractindextext{%
\begin{abstract}
In time-series analysis, nonlinear temporal misalignment remains a pivotal challenge that forestalls even
simple averaging. Since its introduction, the Diffeomorphic Temporal Alignment Net (DTAN), which we first introduced in~\cite{Shapira:NIPS:2019:DTAN} and further developed in~\cite{Shapira:ICML:2023:RFDTAN}, 
has proven itself as an effective solution for this problem (the conference papers~\cite{Shapira:NIPS:2019:DTAN} and~\cite{Shapira:ICML:2023:RFDTAN} are earlier partial versions of the current manuscript). DTAN predicts and applies diffeomorphic transformations in an input-dependent manner, thus facilitating the joint alignment (JA) and averaging of time-series ensembles in an unsupervised or a weakly-supervised manner. The inherent challenges of the weakly/unsupervised setting, particularly the risk of trivial solutions through excessive signal distortion, are mitigated using either one of two distinct strategies: 1) a regularization term for warps; 2) using the Inverse Consistency Averaging Error (ICAE).
The latter is a novel, regularization-free approach which also facilitates the JA of variable-length signals.
We also further extend our framework to incorporate multi-task learning (MT-DTAN), enabling simultaneous time-series alignment and classification. Additionally, we conduct a comprehensive evaluation of different backbone architectures, demonstrating their efficacy in time-series alignment tasks. Finally, we showcase the utility of our approach in enabling Principal Component Analysis (PCA) for misaligned time-series data. Extensive experiments across 128 UCR datasets validate the superiority of our approach over contemporary averaging methods, including both traditional and learning-based approaches, marking a significant advancement in the field of time-series analysis.
\end{abstract}
}

\maketitle

\IEEEdisplaynontitleabstractindextext
\IEEEpeerreviewmaketitle

\IEEEraisesectionheading{\section{Introduction}\label{Sec:Introduction}}

Time-series data often exhibits a significant amount of misalignment (also known as nonlinear time warping);
i.e., typically the observations are
\begin{align}\label{Eqn:MisalignedDataFwd}
  (u_i)_{i=1}^N = (v_i \circ w_i)_{i=1}^N 
\end{align}
where
$u_i$ is the $i^\mathrm{th}$ misaligned signal,
$v_i$ is the $i^\mathrm{th}$ latent aligned signal, 
$w_i$ is a latent warp of the domain of $v_i$, 
and $N$ is the number of signals.
For technical reasons, the misalignment is usually viewed 
in terms of $T_i \triangleq w_i^{-1}$, the inverse warp
of $w_i$, implicitly suggesting $w_i$ is invertible.
It is also typically assumed that 
$(T_i)_{i=1}^N$ belong to some nominal family of warps, parametrized by
$\btheta$:
 \begin{align}\label{Eqn:MisalignedDataInv}
 (v_i)_{i=1}^N = (u_i\circ {T^{\btheta_i}})_{i=1}^N  \,, \quad  T_i=T^{\btheta_i}\in\Tcal\; \forall i\in(1,\ldots,N)
\, .
\end{align}
The nuisance warps, $(T^{\btheta_i})_{i=1}^N$, 
create a fictitious variability in the range of the signals, 
confounding their statistical analysis. 

\begin{figure}[t]
\centering
\def\figwidth{0.98\linewidth } 


\begin{subfigure}{\figwidth}
 \centering
{\includegraphics[trim = 6mm 2mm 7mm 0mm, clip, width=.95\linewidth]{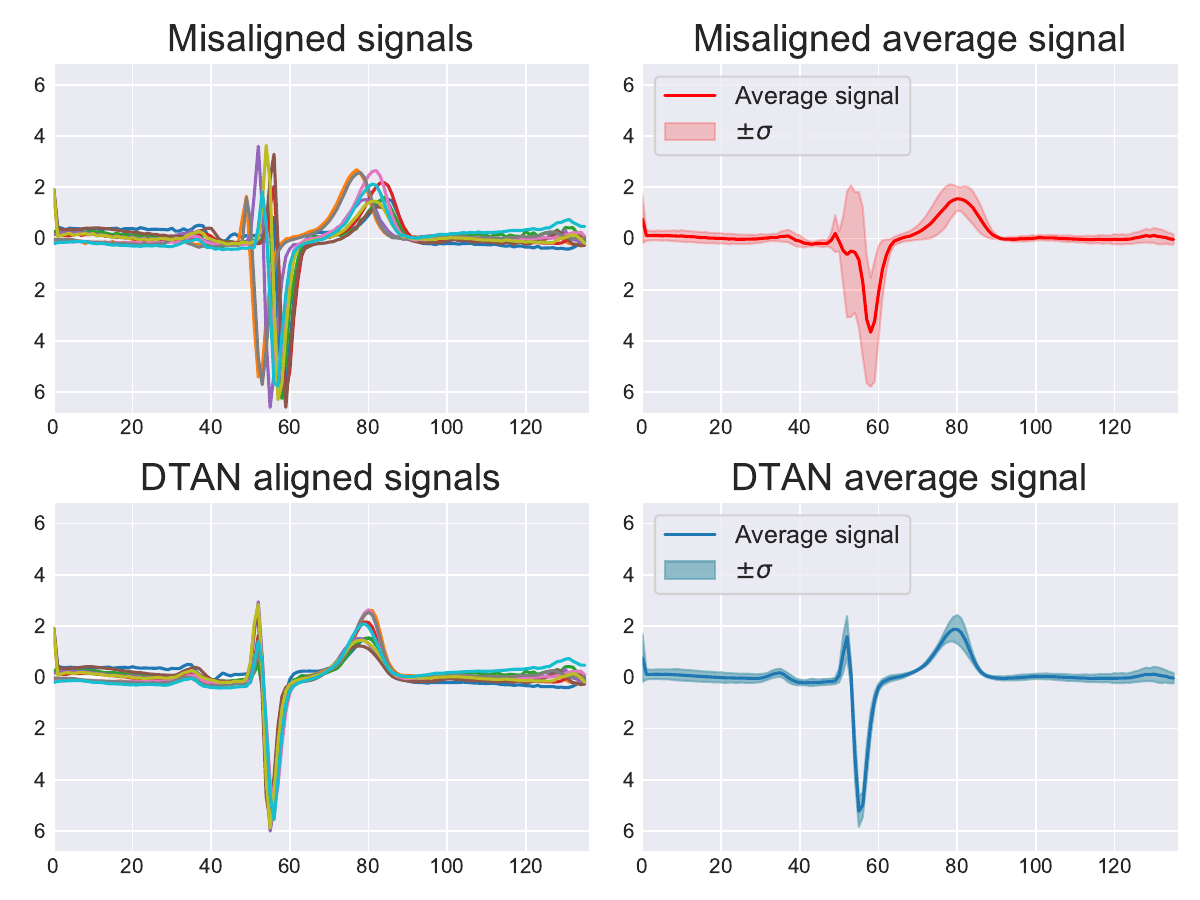}}
\end{subfigure}

\caption{An illustration of the joint-alignment problem in ECG data. 
The data shown is test data. 
    Top: temporal misalignment between ECG signals and its effect on the sample mean (\textit{ECGFiveDays} Dataset).
    Bottom: joint-alignment prediction by DTAN at test time.}
\label{fig:intro}
\end{figure}

To fix ideas, consider ECG recordings from healthy patients during rest. Suppose that the signals
were partitioned correctly such that each segment corresponds to a heartbeat, and that these segments were resampled 
to have equal length (\eg, see~\autoref{fig:intro}). Each resampled segment is then 
viewed as a distinct signal.
The sample mean of these usually-misaligned signals (even when restricted to single-patient recordings) would not look like
the iconic
ECG sinus rhythm; rather, it would smear the correct peaks and valleys and/or contain superfluous ones. This is unfortunate as the sample mean 
has numerous applications in data analysis.  

Moreover, even if one succeeds somehow
in aligning a currently-available recording batch,
upon the arrival of new data batches, the latter will also need to be aligned; \ie, one would like to generalize the inferred alignment from the original batch to the new data
without having to solve a new optimization problem. This is especially the case if the new dataset is much larger than the original one; \eg, imagine a hospital solving the problem once,
and then generalizing its solution, essentially at no cost, to align all the data 
collected in the following year. 
Finally,
these issues become even more critical for multi-class data (\eg, healthy/sick patients), where only in the original batch
we know which signal belongs to which class; \ie, seemingly, the new data will have to be
explicitly classified before its within-class alignment. 

\begin{figure}[t]
  \begin{tikzpicture}
  
  \definecolor{darkgray176}{RGB}{176,176,176}
  \definecolor{darkorange25512714}{RGB}{255,127,14}
  \definecolor{lightgray204}{RGB}{204,204,204}
  \definecolor{steelblue31119180}{RGB}{31,119,180}
  
\begin{axis}[
    legend cell align={left},
    legend style={
        fill opacity=0.8,
        draw opacity=1,
        text opacity=1,
        at={(0.58,0.03)},
        anchor=south west,
        draw=lightgray204,
        font=\scriptsize 
    },
    tick align=outside,
    tick pos=left,
    x grid style={darkgray176},
    xlabel={\# Principal components},
    xlabel style={font=\scriptsize}, 
    xmajorgrids,
    xmin=0.55, xmax=10.45,
    xtick={1, 2, 3, 4, 5, 6, 7, 8, 9, 10},
    xticklabel style={font=\scriptsize}, 
    xtick style={color=black},
    y grid style={darkgray176},
    ylabel={\% of explained variance},
    ylabel style={font=\scriptsize, xshift=-0.5mm}, 
    ymajorgrids,
    ymin=0.6, ymax=1.02,
    ytick={0.5, 0.6, 0.7, 0.8, 0.9, 1.0},
    yticklabel style={font=\scriptsize}, 
    ytick style={color=black},
    width=0.99\textwidth, height=0.5\textwidth,
]
  \addplot [semithick, steelblue31119180, const plot mark right, mark=asterisk, mark size=3, mark options={solid}]
  table {%
  1 0.66707855463028
  2 0.847690880298615
  3 0.888672709465027
  4 0.913015007972717
  5 0.929820597171783
  6 0.940702140331268
  7 0.948945224285126
  8 0.956215500831604
  9 0.963076233863831
  10 0.96934300661087
  };
  \addlegendentry{Original data}
  \addplot [semithick, darkorange25512714, const plot mark right, mark=asterisk, mark size=3, mark options={solid}]
  table {%
  1 0.944242119789124
  2 0.974372923374176
  3 0.991641521453857
  4 0.993819355964661
  5 0.995397627353668
  6 0.996692180633545
  7 0.997215807437897
  8 0.997702896595001
  9 0.997989892959595
  10 0.99823135137558
  };
  \addlegendentry{Aligned data}
  \end{axis}
  
  \end{tikzpicture}
  \label{fig:pca:variance}
    \centering
    \def\figwidth{0.92\linewidth} 

    \begin{tikzpicture}
        \node[inner sep=0] (img1) at (0,0.2) {
            \includegraphics[trim = 1mm 1mm 1mm 10mm, clip, width=\figwidth]{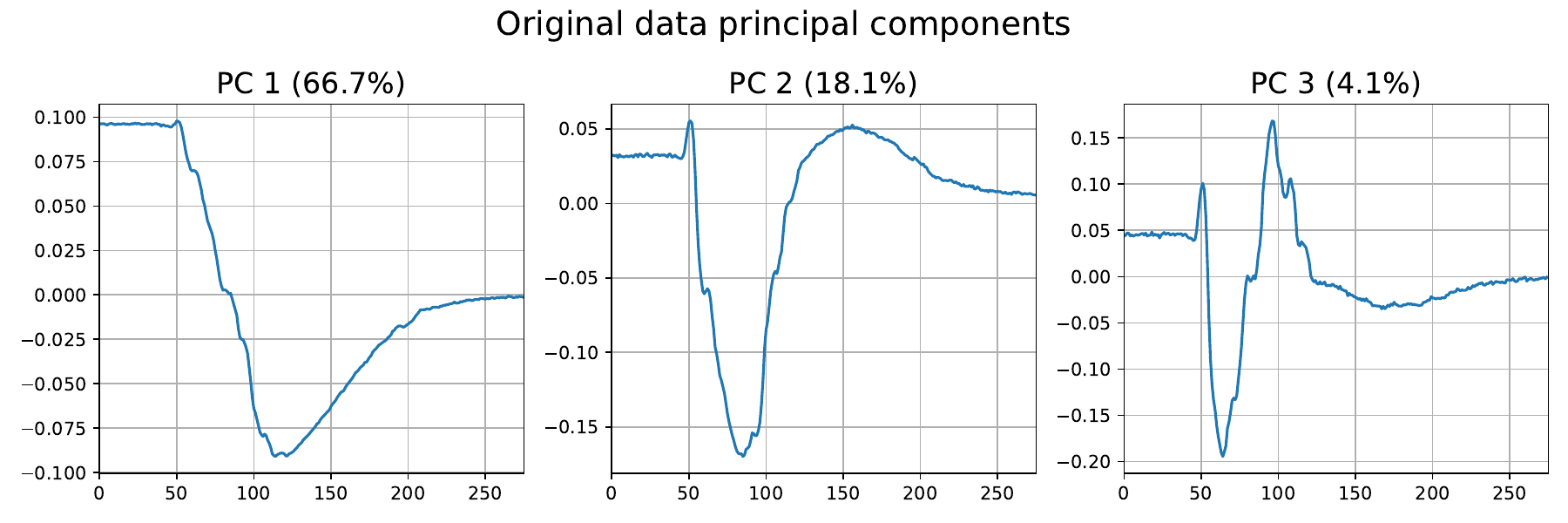}
        };
        \node[rotate=90, left=1mm of img1, anchor=center, font=\scriptsize] {Original PCs};

        \node[inner sep=0, below=-1mm of img1] (img2) {
            \includegraphics[trim = 1mm 1mm 1mm 10mm, clip, width=\figwidth]{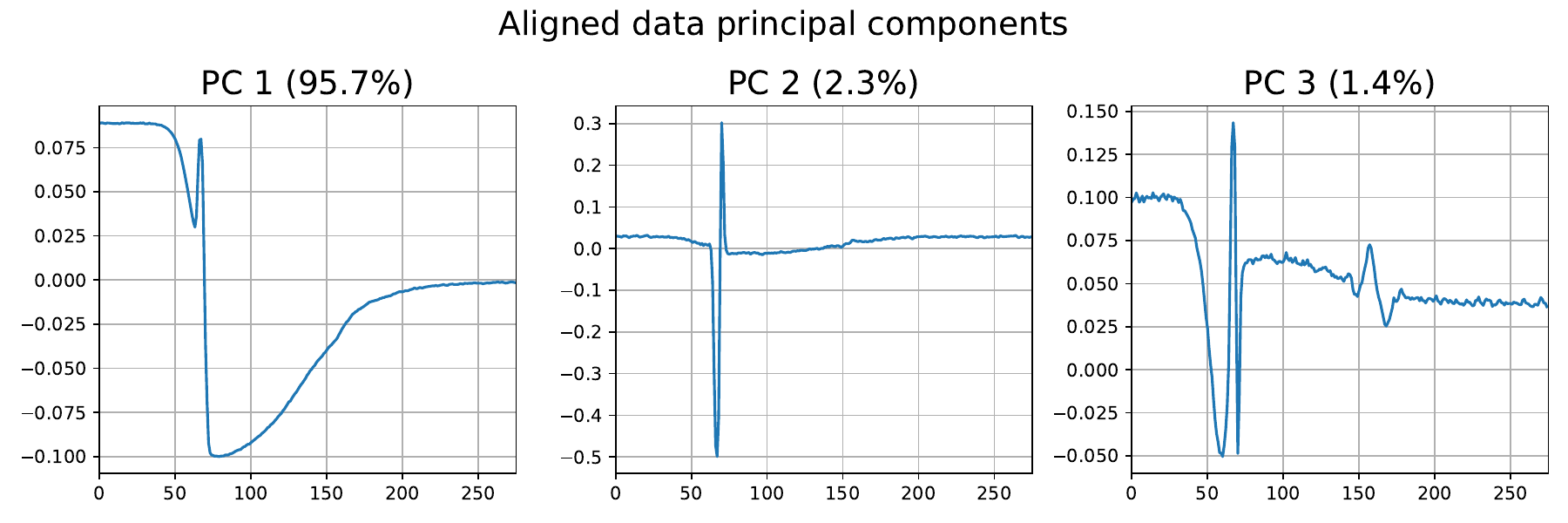}
        };
        \node[rotate=90, left=1mm of img2, anchor=center, font=\scriptsize] {Aligned PCs};

        \node[inner sep=0, below=0mm of img2] (img3) {
            \includegraphics[trim = 1mm 1mm 1mm 1mm, clip, width=\figwidth]{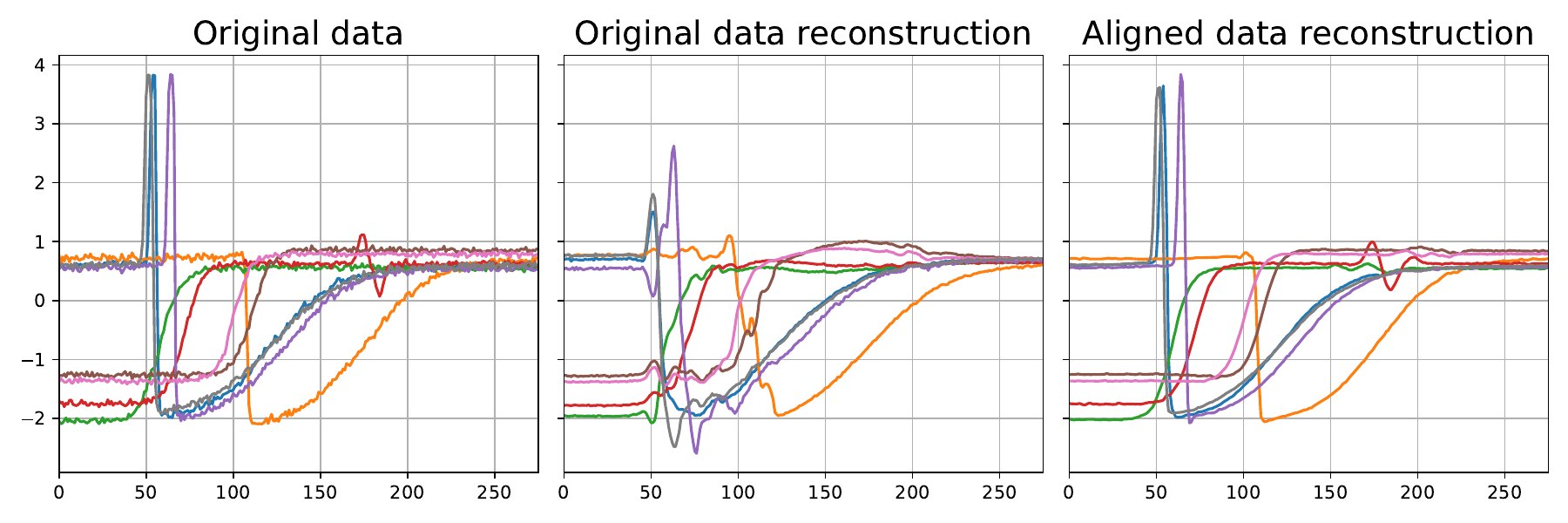}
        };
        \node[rotate=90, left=1mm of img3, anchor=center, font=\scriptsize] {Reconstruction};
    \end{tikzpicture}
    
    \caption{The benefits of joint alignment for dimensionality reduction, evaluated on the \textit{Trace} dataset~\cite{Dau:2019:ucr} using Principal Component Analysis. The top panel shows the cumulative explained variance as a function of the number of Principal Components (PCs). The middle-top and middle-bottom panels depict the first 3 PCs of the original and DTAN-aligned data, respectively. The bottom panel illustrates the reconstruction of the original and (inverse warped) aligned data using the first 6 PCs.}
    
    \label{fig:pca}
    \end{figure}
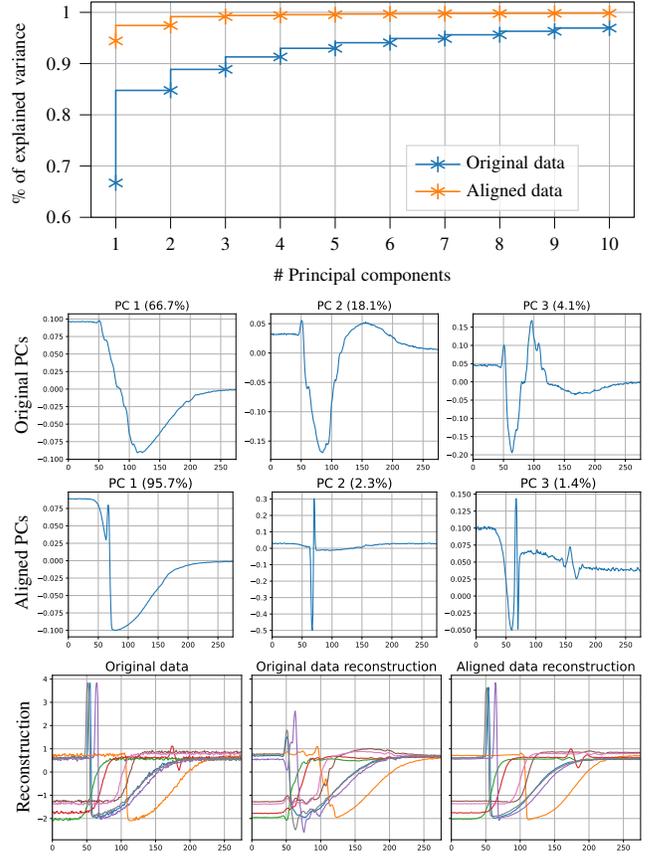

To further contextualize this concept, let us consider the application of dimensionality reduction in time-series data, using Principal Component Analysis (PCA) as an example. PCA is designed to identify the Principal Components (PCs) that capture the maximum variance in a dataset. For time series, the initial step in PCA involves centering the data by subtracting the mean sequence from each signal. 
However, if the sequences are misaligned, this mean might not accurately represent the true underlying structure. This, in turn, will lead to 
more variance at each time step, which will subsequently require more PCs to explain the data. 
Thus, unwarping the signals will allow for fewer PCs to be needed to effectively describe the data, as they are no longer compensating for the distortions caused by misalignment (see~\autoref{fig:pca} for an illustration).

A popular attempt to solve the problem relies on \textbf{pairwise alignments}. 
 Let $u_i=(u_i(t))_{t=1}^n$ and $u_j=(u_j(t))_{t=1}^m$ be two real-valued discrete-time signals of lengths $n$ and $m$, respectively. 
 The optimal pairwise alignment of $u_j$ towards $u_i$, under some dissimilarity  measure $D$, is defined by 
 \begin{align}\label{Eq:pairwise}
     T^* = \argmin{T\in\Tcal}D(u_i, u_j\circ T)
 \end{align}
 where $\circ$ denotes function composition and $\Tcal$ is a family of \emph{warps} (or warping functions); namely, every $T\in\Tcal$ 
 is a function  $T:\Omega \to \RR$ where $\Omega\subset \RR $ is an interval containing $\set{1,\ldots,m}$. 
 For instance, Dynamic Time Warping (DTW) provides the optimal discrete warping path between the time indices of $u_i$ and $u_j$ via dynamic programming, where $D$ is (usually) a Euclidean distance~\cite{Sakoe:ICA:1971:DTW1}.
 More generally, while $u_i$ and $u_j$ are defined over discrete domains (\ie, $\set{1,\ldots,n}$ and $\set{1,\ldots,m}$), 
 the notation $u_j\circ T$ in \autoref{Eq:pairwise} implicitly assumes that the value of  $u_j(t')$ at every $t'\in \RR$
 is determined, using interpolation techniques, from (possibly a subset of) the $m$ given values, $(u_j(t))_{t=1}^m$.

  In this paper, which extends our two conference papers~\cite{Shapira:NIPS:2019:DTAN, Shapira:ICML:2023:RFDTAN}, we focus on continuously-defined warps that are 
 order-preserving \emph{diffeomorphisms}. 
A diffeomorphism (namely, a differentiable invertible function whose inverse is differentiable),
 is a natural choice for representing time warping~\cite{Mumford:Book:2010:PT}. 
Since spaces of diffeomorphisms are large, and to discourage unfavorable solutions, typically a regularization 
term, denoted by $T\mapsto\Rcal(T;\lambda)$ and parameterized by \emph{hyperparameters} (HP), $\lambda$, is added to the objective function; \eg, $\Rcal$ might penalize lack of smoothness (in the machine-learning sense, not calculus) or large deviations from the identity map.
Hence, \autoref{Eq:pairwise} is commonly replaced with 
  \begin{align}
     T^* = \argmin{T\in\Tcal}D(u_i, u_j\circ T)+\Rcal(T;\lambda)
 \end{align}\label{Eq:pairwiseWithReg}
 where $\Tcal$ is a space of 1D diffeomoprhisms from $\Omega$ into $\RR$. 


In the case of an ensemble of $N$ signals, $(u_i)_{i=1}^N$ where $N>2$, the pairwise approach usually does not generalize well,
is prone to drift errors, and might introduce inconsistent solutions. 
This motivates approaches for \textbf{joint alignment} (JA),
also known as global alignment or multiple-sequence alignment.
The JA problem is often formulated as
\begin{align}\label{eq:JA}
    (T_i^*)_{i=1}^N,\mu   = \argmin{(T_i)_{i=1}^N\in\Tcal, u } \sum\limits_{i=1}^N 
     D(u , u_i\circ T_i)+ \Rcal(T_i;\lambda) \,    
\end{align}
where $\Tcal$, $\Rcal(\cdot;\lambda)$, and $D$ are as before, 
 $T_i$ is the latent warp associated with $u_i$, and $\mu $ is a latent signal, conceptually thought of as the \emph{average signal} (or \emph{centroid}) of the ensemble. This optimization task may also be amortized
via the training of a deep net (\eg,~\cite{Shapira:NIPS:2019:DTAN,huang:2021:residual,Martinez:ICML:2022:closed,Shapira:ICML:2023:RFDTAN}).

\begin{figure*}[t]
\begin{center}
\centering
\def\figwidth{0.99\linewidth}

\includegraphics[trim =0mm 6mm 0mm 0mm, clip, width=\figwidth]{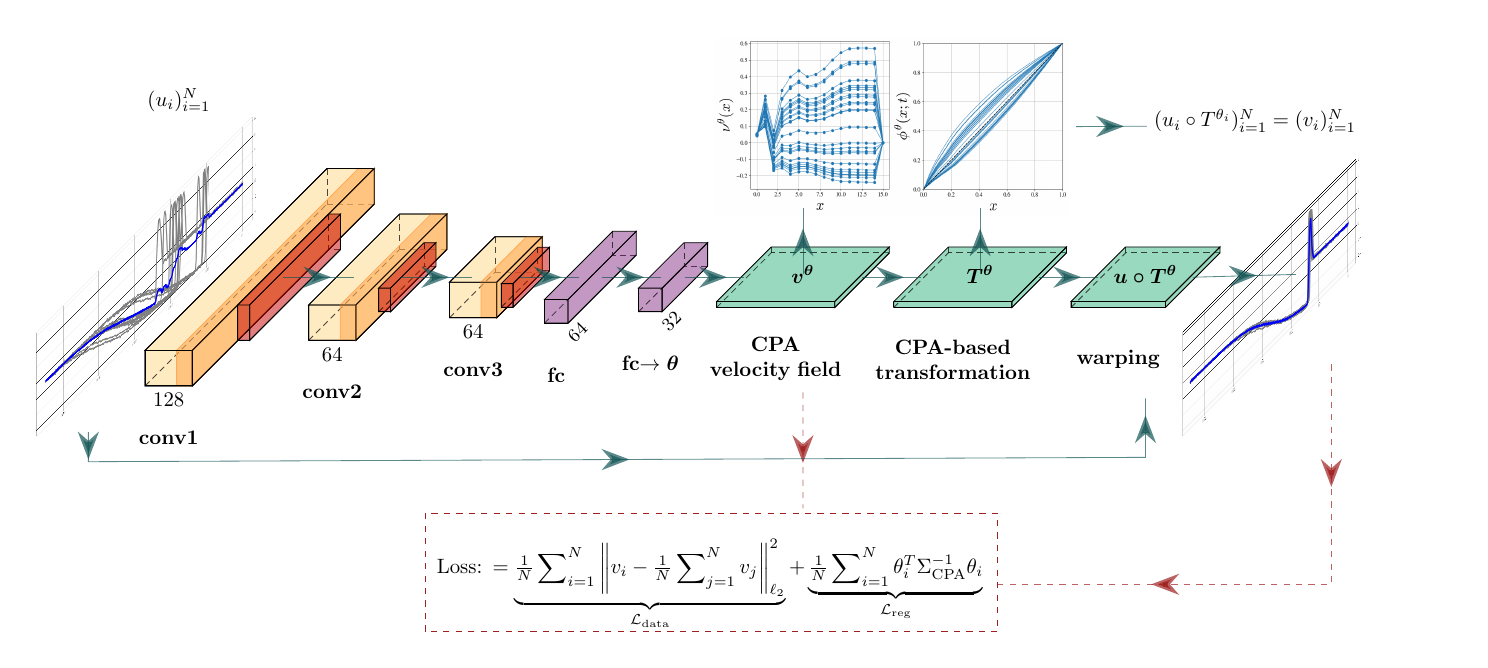}

\caption{DTAN joint alignment demonstrated on a class of the ``Trace" dataset~\cite{Chen:UCR:Archive:2015} with a simple 1D ConvNet backbone which was used in~\cite{Shapira:NIPS:2019:DTAN}.
Signals are denoted in gray and their average in blue. 
Each Convolution layer is followed by a ReLU, Batch Normalization, and a Max-Pooling layer. The final Fully-Connected 
layer (fc) predicts the warping parameters, $\btheta$, of the CPA velocity fields, $v^{\btheta}$, which is then integrated to form a CPAB warp, $T^{\btheta}$. The latter, in turn, is applied to the input signal ($u$) to create the output, $u\circ T^{\btheta}=v$. The loss consists of the empirical within-class variance ($\Lcal_{\mathrm{data}}$) and a regularization term on $\btheta$ ($\Lcal_{\mathrm{reg}}$).}

\label{fig:fig_main}
\end{center}
\end{figure*}

\emph{We argue that this problem should be seen as a learning one,
mostly due to the need for generalization}.
Particularly, we propose a novel deep-learning (DL) approach for the joint alignment of time-series data.
More specifically, inspired by computer-vision 
and/or pattern-theoretic solutions for misaligned images
(\eg, congealing~\cite{Miller:CVPR:2000:learning,Learned:PAMI:2006:align,Huang:CVPR:2007:unsupervised,Huang:NIPS:2012:learning,Cox:CVPR:2008:LS,Cox:ICCV:2009:LS}, 
efficient diffeomorphisms~\cite{Freifeld:ICCV:2015:CPAB,Freifeld:PAMI:2017:CPAB, Zhang:IPMI:2015,Zhang:IJCV:2018:fast}, and spatial transformer nets~\cite{Jaderberg:NIPS:2015:spatial,Lin:CPVR:2017:inverse,Skafte:CVPR:2018:DDTN}),
we introduce the Diffeomorphic Temporal Alignment Network (DTAN) which learns an input-dependent diffeomorphic 
time warping to its input signal to minimize a joint-alignment loss (see~\autoref{fig:fig_main} for a detailed illustration of the proposed model). The diffeomorphism family we use, called CPAB
\cite{Freifeld:ICCV:2015:CPAB,Freifeld:PAMI:2017:CPAB}, is 
based on the integration of piecewise affine velocity fields 
 and will be further discussed in~\autoref{cpab}.
In the single-class case, DTAN is completely unsupervised. For multi-class problems, 
we propose a weakly-supervised method that results in a single model (for all classes) that learns how to perform within-class joint alignment.
We demonstrate the utility of the proposed framework on real-world datasets with applications to time-series
 joint alignment, averaging, classification, and dimensionality reduction. 
 
Below we list our 6 key contributions. Contributions 1-2-3 appeared  in~\cite{Shapira:NIPS:2019:DTAN,Shapira:ICML:2023:RFDTAN}) 
while contributions 4-5-6 are new. 
\begin{enumerate}[leftmargin=.5cm]
\item DTAN, a DL framework for learning time series joint alignment and averaging~\cite{Shapira:NIPS:2019:DTAN}.
\item RDTAN: a recurrent version of DTAN which predicts diffeomorphisms derived from non-stationary velocity fields~\cite{Shapira:NIPS:2019:DTAN}.
\item A regularization-free objective function for the JA task - the \emph{Inverse-Consistency Averaging Error} (ICAE)~\cite{Shapira:ICML:2023:RFDTAN}.
\item DTAN-MT: a multi-task version of DTAN that learns both time-series alignment and classification, resulting in
better separation between the aligned classes.
\item Evaluation of prominent time-series classification DL architectures as the backbone of the Temporal Transformer module.
\item Analyzing DTAN's effect on dimensionality reduction via PCA.
\end{enumerate}
 We conclude this section with a timely remark: throughout the paper, the term ``transformer net" should be  understood in the same sense it was used in STN~\cite{Jaderberg:NIPS:2015:spatial} 
 or TTN~\cite{Shapira:NIPS:2019:DTAN} (\ie, a Spatial or Tempiral Transform Net), and is not to be confused with how it is used in Natural Language Processing or Vision Transformers. 

\begin{table}[t]
\newcommand{\cmark}{\ding{51}}%
\newcommand{\xmark}{\ding{55}}%
\caption{Comparing JA/averaging methods. Learning gives the ability to generalize JA to new data. 
VL indicates whether the method supports variable-length signals.}
\label{table:methods}
\vskip 0.15in
\begin{center}
\begin{small}
\begin{sc}
\scalebox{0.83}{
\begin{tabular}{lccccr}
\toprule
Method & Reg.-free &Optimization & Learning & VL\\
\midrule
Euclidean & \cmark& \textit{N/A} & \xmark & \cmark \\
DBA     & \cmark&  EM&  \xmark & \cmark \\
SoftDTW & \xmark& L-BFGS & \xmark & \cmark \\
DTAN w/ WCSS & \xmark &   DL training    & \cmark & \xmark \\
DTAN w/ $\Lcal_{\mathrm{ICAE}}$ & \cmark    &DL training    & \cmark & \cmark \\
\bottomrule
\end{tabular}
}
\end{sc}
\end{small}
\end{center}
\vskip -0.1in
\end{table}

\section{Related Work}\label{Sec:previous}
\textbf{Dynamic Time Warping (DTW)} is a popular distance measure (or discrepancy) between a time-series pair~\cite{Sakoe:ICA:1971:DTW1,Sakoe:ASSP:1971:DTW2}. Given two signals of lengths $n$ and $m$, DTW computes
the best discrete alignment path in the $n\times m$ pairwise distance matrix. While its complexity is $O(nm)$, enforcing certain constraints on DTW results in a linear complexity. 
However, generalizing DTW from the pairwise case to the JA of multiple signals is prohibitively expensive
since the complexity of finding the optimal discrete alignment between $N$ signals of length $n$ is $O(n^N)$.
To overcome this limitation, several JA methods, working under the DTW geometry, were proposed.
The DTW-Barycenter Averaging (DBA)~\cite{Petitjean:2011:global,Petitjean:2014:dynamic} employs an Expectation-Maximization (EM) approach 
to refine a signal that minimizes the sum of DTW distances from the data; \ie, 
it alternates between finding  $\mu$ (while fixing $(T_i)_{i=1}^N$),
\begin{align}\label{eq:ts:average}
    \mu = \argmin{u}\sum\limits_{i=1}^N D(u, u_i\circ T_i) \, ,
\end{align}
and finding  discretely-defined $(T_i)_{i=1}^N$ (while fixing $\mu$),
\begin{align}\label{eq:loss:and:prior}
    (T_i^*)_{i=1}^N = \argmin{(T_i)_{i=1}^N\in\Tcal} \sum\limits_{i=1}^N 
     D(\mu, u_i\circ T_i) \, .
\end{align}

  SoftDTW~\cite{cuturi:2017:soft}, a soft-minimum variant of DTW, extends DBA. 
  Instead of using EM, SoftDBA computes $\mu$ via gradient-based optimization. SoftDTW has one HP, $\gamma$, that controls the smoothness of the alignment ($\gamma=0$
  gives the original DTW score). 
  SoftDTW-divergence~\cite{Blondel:2021:differentiable} modifies SoftDTW to a proper positive-definite divergence. Both of these optimization-based methods \emph{do not learn} how to find the JA of \emph{new} data; \ie, 
  when new signals arrive, they must be run from scratch in order to achieve JA of the new ensemble. While it is possible to align the new data to the previously-found $\mu$ in a pairwise manner, this leads to inferior results (see~\autoref{Sec:Results}). 
  Additionally, the time/memory complexity of SoftDTW is $O(mn)$. SoftDTW-div
  has an even worse complexity for a large $n$ or $m$; \eg, results on \texttt{HandOutlines} (the largest UCR dataset in terms of $n\times N$) 
  were not reported by~\cite{Blondel:2021:differentiable}, and when we tried to run SoftDTW (using \texttt{tslearn}~\cite{tavenard:2017:tslearn}) on it, 
  it failed due to memory limitations. 

Other methods include the Global Alignment Kernel (GAK)~\cite{cuturi:2011:gak} on which SoftDTW is based, DTW with Global Invariances
 which generalizes DTW/SoftDTW to both time and space~\cite{Vayer:2020:time}, and 
Neural Time Warping 
that relaxes the original problem
to a continuous optimization using a neural net (albeit limited in the number of signals it can jointly align)~\cite{kawano:ICASSP:2020:neural}. 

\textbf{Spaces of Diffeomorphisms} are often used for modeling warping paths between sequences;  \eg,~\cite{Srivastava:2010:shape,srivastava:2011:registration} proposed differomoprhisms based on the square-root velocity function (SRVF) representation. 
However, the employment of diffeomorphisms in DL used to be hindered by the associated expensive computations and/or approximation/discretization schemes. For example, this is why diffeomorphisms could not initially be used effectively within a
Spatial Transformer Net (STN)~\cite{Jaderberg:NIPS:2015:spatial} since training the latter requires a large number of evaluations of both $x\mapsto T^\btheta(x)$ and $x\mapsto \nabla_\btheta T^\btheta(x)$ (where $\btheta$ 
parameterizes the chosen diffeomorphism family), and these quantities are computed at multiple values of $x$.   
 This has changed, however, with the emergence of new methods~\cite{Skafte:CVPR:2018:DDTN,Balakrishnan:CVPR:2018:UnsupervisedDeformable}. 
In particular, \cite{Skafte:CVPR:2018:DDTN} built on the CPAB diffeomorphisms (see below) 
to propose the first diffeomorphic STNs.

\textbf{CPAB Diffeomorphisms~\cite{Freifeld:ICCV:2015:CPAB,Freifeld:PAMI:2017:CPAB}}. 
The name CPAB, short for CPA-Based, stems from the fact that these parametric diffeomorphisms are based on Continuous Piecewise-Affine (CPA) velocity fields. Of note, in 1D, the CPAB warp, $x\mapsto T^\btheta(x)$, has a closed form~\cite{Freifeld:ICCV:2015:CPAB}. 
The expressiveness and efficiency of the CPAB warps make them an invaluable tool in DL (see, \eg, ~\cite{Hauberg:AISTATS:2016:DA,Skafte:CVPR:2018:DDTN,Skafte:NIPS:2019:explicit,Shapira:NIPS:2019:DTAN,kaufman:icip:2021:cyclic,Shacht:2021:single,Schwobel:2022:UAI:pstn,Martinez:ICML:2022:closed, Neifar:2022l:everaging,Kryeem:ICCV:2023:personalized,Wang:AAAI:2024:Animation,kryeem:CVIU:2025:action,Chelly:ECCV:2024:ditac,Mantri:NIPS:2024:DIGRAF}) and thus this work uses them too.  
However, our method is not limited to this choice of $\Tcal$.

A \textbf{Temporal Transformer Net (TTN)} is the 1D variant of the STN, where the latter is a DL module which, given a transformation family, predicts and applies a transformation to its input for a downstream task. 
\citet{Lohit:CVPR:2019:temporal} use TTNs with discretized diffeomorphisms for learning rate-invariant discriminative warps. 
The SRVF framework was integrated into TTNs to either predict DTW-based warping functions~\cite{nunez:CVPRW:2020:deep}, learn a generative model over the distribution of SRVF warps~\cite{Nunez:2021:srvfnet}, and time-series JA~\cite{Chen:2021:srvfregnet}. However, computations in these nonparametric warps do not scale well with the signal length.

The \textbf{Diffeomorphic Temporal Alignment Net (DTAN)}~\cite{Shapira:NIPS:2019:DTAN}
is a diffeomorphic TTN that, using the parametric and highly-expressive CPAB warps, offers an effective learning-based solution for JA and averaging.~\citet{Shapira:NIPS:2019:DTAN} based their DTAN implementation on \texttt{libcpab}~\cite{Detlefsen:2018:libcpab}.  
Recently,~\citet{Martinez:ICML:2022:closed} released another CPAB library, Diffeomorphic Fast Warping (\texttt{DIFW}), 
which, while being similar to \texttt{libcpab} (and is, in fact, based on it), is even faster, largely due to the smart discovery of a closed-form gradient~\cite{Martinez:ICML:2022:closed} for CPAB warps. 
Together with some other changes and an extensive HP tuning on the test data, 
this let them propose
a DTAN implementation with SOTA results in terms of Nearest Centroid Classification (NCC) accuracy, a standard metric for time-series averaging.  
 Henceforth will refer to the DTAN implementations from~\cite{Shapira:NIPS:2019:DTAN} and~\cite{Martinez:ICML:2022:closed}
 as DTAN$_{\mathrm{libcpab}}$ and DTAN$_{\mathrm{DIFW}}$, respectively. 
Lastly, ResNet-TW~\cite{huang:2021:residual} also predicts CPAB warps albeit via the Large Deformation Diffeomorphic Metric Mapping 
framework~\cite{Beg:IJCV:2005}.

\textbf{Warp Regularization.} As is typical with diffeomorphisms, 
CPAB warps too are usually regularized. 
In particular, the three works above~\cite{Shapira:NIPS:2019:DTAN,huang:2021:residual,Martinez:ICML:2022:closed},
who all use the \textbf{within-class-sum-of-squares (WCSS) loss}, 
 also use the following regularization from~\cite{Freifeld:ICCV:2015:CPAB}, 
$\Rcal(T^{\btheta_i};\lambda) = \btheta_i^{\top}\bSigma^{-1}_{\mathrm{CPA}}\btheta_i$. 
The matrix $\bSigma_{\mathrm{CPA}}$ is the covariance of a
zero-mean Gaussian smoothness prior over CPA velocity fields
and has two HPs: $\lambda_{\bSigma}$, which controls the overall variance, and
$\lambda_{\mathrm{smooth}}$, which controls the smoothness of the fields. Additionally, all these three methods predict a varying number of warps (denoted by $N_{\mathrm{warps}}$), such that their composition yields the final warp.

We conclude the section with ~\autoref{table:methods} that summarizes the differences between several JA/averaging methods and ours.
\section{Preliminaries}\label{Sec:Background}
\subsection{Temporal Transformer Nets} 
Given $\Tcal$ , a differentiable transformation family parameterized by $\btheta$, a Spatial Transformer (ST) layer performs a
 learnable input-dependent warp w.r.t a given objective function~\cite{Jaderberg:NIPS:2015:spatial}.
  Reducing this from images (a 2D domain) to time series (1D), one obtains a Temporal Transformer (TT) layer. A TTN is a neural net with at least one TT layer. In more detail, let $u$ denote the input of the TT layer. Its outputs consist of $\btheta = f_{\mathrm{loc}}(u)$ and
$v=u\circ T^{\btheta}$, where
$T^{\btheta} \in \Tcal$ is a 1D warp parameterized by $\btheta$. The function $f_{\mathrm{loc}}: u \mapsto \btheta$ is itself a
 neural net called the localization net. Let $\bw$ denote the parameters (also known as weights) of $f_{\mathrm{loc}}$ and let
\begin{align}
  \Lcal((u_i,\btheta_i(u_i;\bw))_{i=1}^N)
\end{align}
denote a loss function. Recall that, as usual, the back-propagation algorithm requires certain partial derivatives and note that one of these derivatives, $\nabla_\btheta (T^{\btheta}( \cdot))$, depends on the choice of $\Tcal$.

The TTN consists of 3 modules:
\begin{enumerate}[leftmargin=.5cm]
 \item \textbf{Localization network.} For an input signal, $u$, the localization network, $f_{\mathrm{loc}}$, predicts the warp's parameters; \ie, $f_{\mathrm{loc}}(u)=\btheta$. 
 Any form of neural network architecture can be used for  $f_{\mathrm{loc}}$, 
 as long as the output layer has $d$ 
 neurons, where $d=\dim(\btheta)$.
 \item \textbf{Parameterized grid generator}. This generator creates a discrete 1D grid 
 of length $M$ (where $M$ is the length of the signals):
 $G=\tuple{x_m}_{m=1}^M\subset[-1,1]$ of evenly-spaced points which are later transformed by the parametrized warp, $T^\btheta$.
 \item \textbf{Differentiable time-series resampler}. The output signal, $v$, is computed 
 by interpolating the values of $v$ at $T^\btheta(G)$ from $u$.
Let $x_{i,m}^{\mathrm{new}} = T^{\btheta_i}(x_m)$ and write the discrete-time $i$-th aligned signal
as
\begin{align}
 v_i &= (v_{i,m})_{m=1}^M=(v_{i,1},\ldots,v_{i,M})\, .
\end{align} 
Note that due to the need to resample the  signal,
rather than having $v_i = u_i \circ T^{\btheta_i}$, we need to also account for the resampling kernel. 
For the popular linear kernel, which is the one used in our work, we obtain (based on~\cite{Jaderberg:NIPS:2015:spatial}),
\begin{align}
   v_{i,m} &= \sum_{m'=1}^M u_{i,m'} \max(0, 1 - |x_{i,m}^{\mathrm{new}}  -m'|) \, .
\end{align}
\end{enumerate}
To propagate the loss to $f_{\mathrm{loc}}$, the resampling kernel must be differentiable, 
which is the case for the linear kernel:
 \begin{align}
   \frac{\partial v_{i,m}}{\partial u_{i,m'}  }  &= \sum_{m'=1}^M \max(0, 1 - |p_{i,m}^{\mathrm{warped}}  -m'|) \\ 
     \frac{\partial v_{i,m}}{\partial (p_{i,m}^{\mathrm{warped}})  }  &= 
     \sum_{m'=1}^M u_{i,m'} 
     \left\{
     \begin{matrix}
 0 & \text{if }& |m' - p_{i,m}^{\mathrm{warped}}| \ge 1  \\
 1 & \text{if }& m' \ge p_{i,m}^{\mathrm{warped}}  \\
 -1& \text{if }& m' < p_{i,m}^{\mathrm{warped}}
     \end{matrix}
     \right.
     \, .
 \end{align}
Here $v_{i,m}$ is the $i^{th}$ warped signal at time point $m$, $u_{i,m'}$ is the input signal at time point $m'$ and $p_{i,m}^{\mathrm{warped}}$ is the $m^{\text{th}}$ point of the sampling grid.
The generalization of these results to multichannel time series is straightforward and thus omitted. 
In~\autoref{cpab} we will specify $\Tcal$ and will discuss its associated derivative, $\nabla_\btheta (T^{\btheta}( \cdot))$.\label{sec:TTN}
\subsection{Deep-learning Time-series Architectures}
The core module of the TTN is the localization network, $f_{\mathrm{loc}}$, which predicts the transformation parameters $\btheta$.
While in~\cite{Shapira:NIPS:2019:DTAN} we have used a simple Temporal Convolutional Neural Network (TCN), here we explore several other 
architectures.

Similar to Computer Vision and Natural Language Processing, the field of Time Series Classification (TSC) also saw 
a recent surge in DL-based classifiers. 
Traditionally, Recurrent Neural Network (RNN)~\cite{Husken:neuro:2003:recurrent,Mikolov:inter:2010:recurrent} were the 
go-to models for this task, as their time-dependent representation allowed them to capture temporal dependencies within signals.
However, leveraging the expressive power of deeper and more recent architectures allowed TCNs to outperform RNNs while offering 
a more efficient training procedure. Fawaz et al., (2018)~\cite{fawaz:2018:deep} provided an extensive review of such architectures
for the TSC task. Their findings pointed to two architectures: Fully-Convolutional Networks (FCN)~\cite{wang:2017:time} and a 1D variant of the
now quintessential Residual Network (ResNet)~\cite{He:ECCV:2016:resnet}.

Another TCN-based architecture is InceptionTime~\cite{Ismail:2020:inceptiontime}. It is a 1D variant of the Inception Network-v4~\cite{Szegedy:AAAI:2017:inception} and is composed of
 an ensemble of Inception modules.
Arguing that frequency information is lost in current TSC models, a wavelet-based neural network structure called multilevel Wavelet
Decomposition Network (mWDN) was proposed~\cite{wang:SIGKDD:multilevel:2018}. It preserves
the advantage of multilevel discrete wavelet decomposition in frequency learning while still enabling the fine-tuning of learnable parameters via 
 back-propagation. The model takes all or partial mWDN decomposed sub-series
  different frequencies as input features and updates its parameters globally for a downstream classification or a forecasting task. As different 
 features at different frequencies are used, the authors coined the integration of mWDN features and a deep-classifier, $\psi(\cdot)$, as 
 \emph{Residual Classification Flow (RCF)}. Here, we utilize the mWDN framework for times-series alignment, which could be thought as 
 \emph{Residual Alignment Flow (RAF)}. In our experiments, $\psi(\cdot)$ is an \emph{InceptionTime} module.
 
Given these recent advancements in DL for TSC, we explore the effect of the aforementioned architectures as 
the ``backbones" of the localization networks in the TTN. In~\autoref{Sec:Results} we provide an evaluation of TCN, 
RNN-FCN, mWDN, and, InceptionTime for time-series joint alignment and averaging under the DTAN framework.
 \label{sec:DLTS}
\begin{figure}[t]
    \centering
    \def\figwidth{0.32\linewidth } 
    \begin{subfigure}{\figwidth}
     \centering
    {\includegraphics[trim = 3.5mm 3mm 4mm 2mm, clip, width=.95\linewidth]{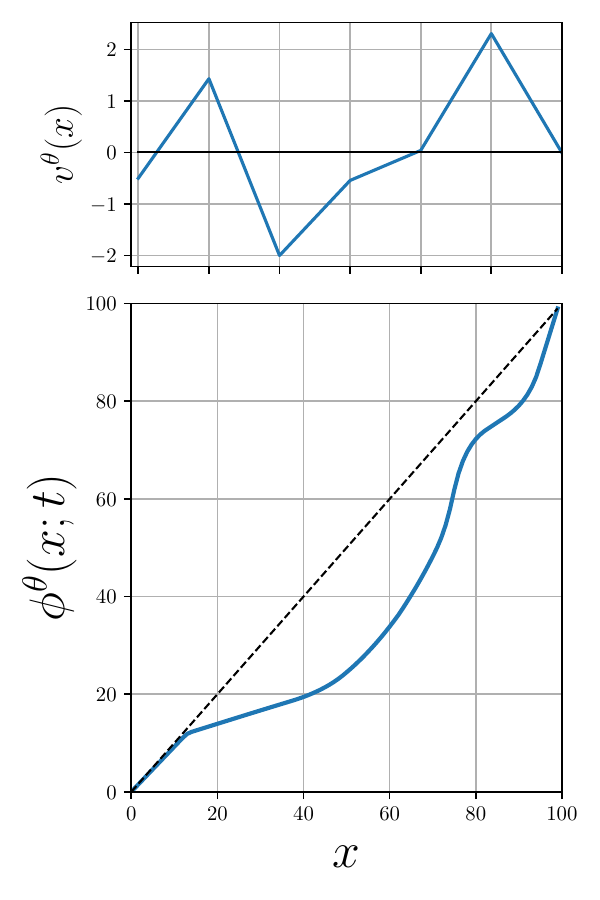}\label{fig:cpab:a}}
    \end{subfigure}
    \begin{subfigure}{\figwidth}
     \centering
    {\includegraphics[trim = 4mm 3mm 4mm 2mm, clip, width=.95\linewidth]{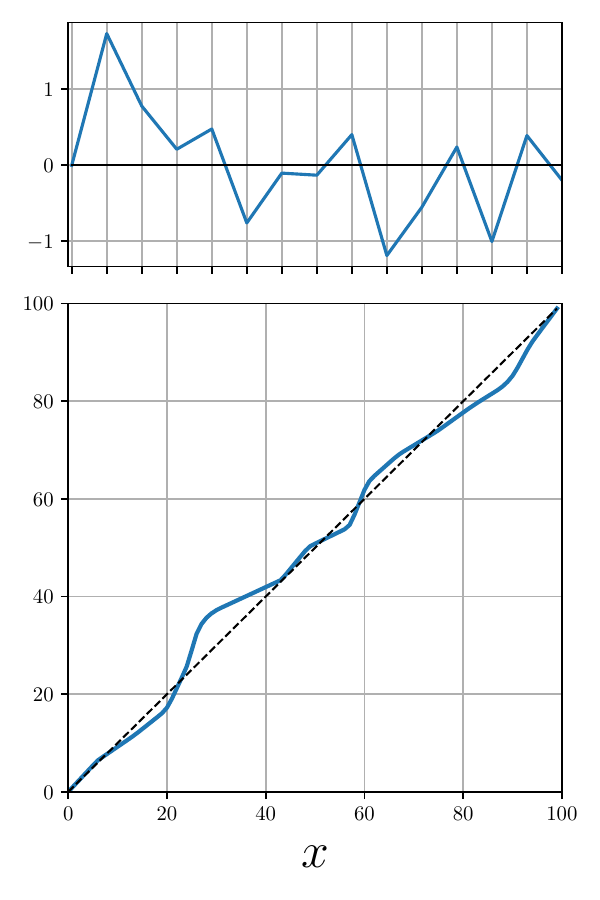}\label{fig:cpab:b}}
    \end{subfigure}
    \begin{subfigure}{\figwidth}
        \centering
       {\includegraphics[trim = 4mm 3mm 4mm 2mm, clip, width=.95\linewidth]{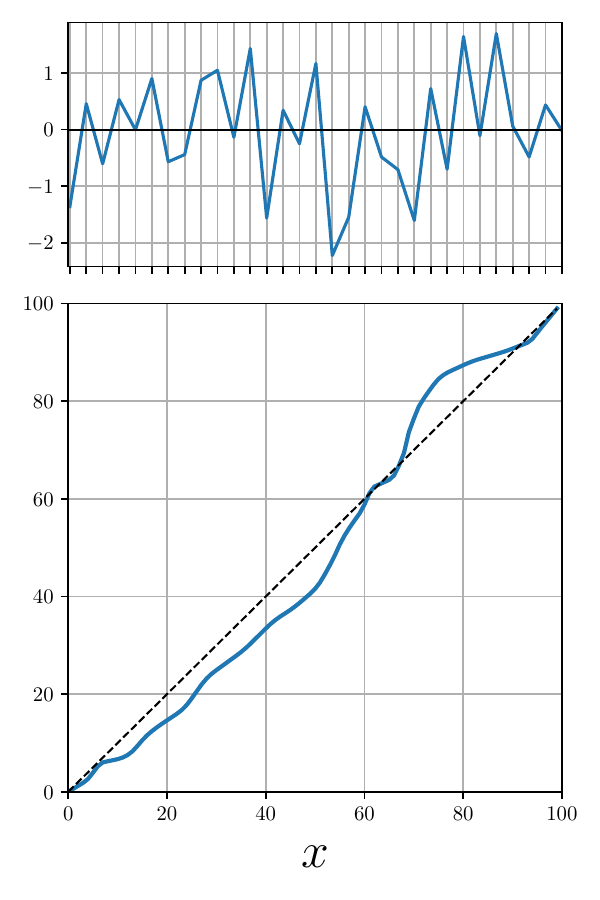}\label{fig:cpab:c}}
    \end{subfigure}
    
    \caption{
    CPAB warps for different partitions of $\Omega\in{\{8, 16,32\}}$.
     Top: Continuous Piecewise-Affine (CPA) velocity fields.
    Bottom: The resulting CPAB warp, obtained via integration of $v^\btheta$.
    }
    \label{fig:cpab1}
    \end{figure}

\begin{figure}[t]
    \centering
    \def\figwidth{0.48\linewidth } 
    \begin{subfigure}{\figwidth}
     \centering
    {\includegraphics[trim = 3mm 2mm 4mm 4mm, clip, width=.95\linewidth]{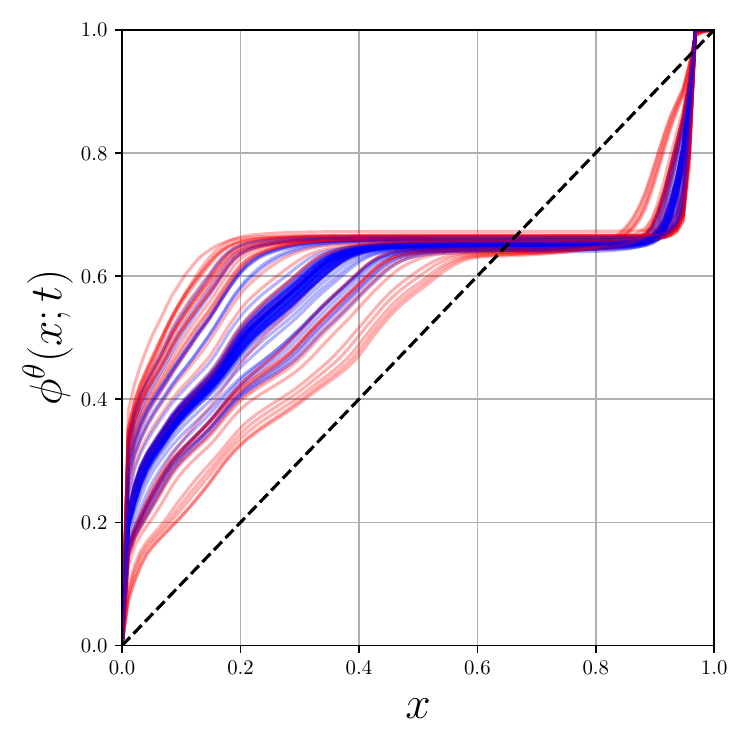}\label{fig:prior:a}}
    \end{subfigure}
    \begin{subfigure}{\figwidth}
     \centering
    {\includegraphics[trim = 3mm 2mm 4mm 4mm, clip, width=.95\linewidth]{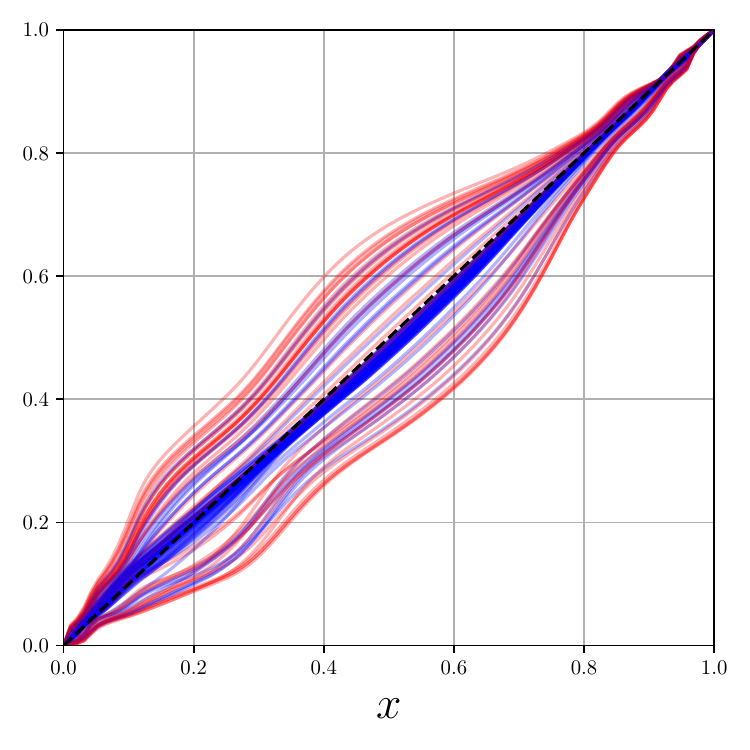}\label{fig:prior:b}}
    \end{subfigure}
    
    \caption{
    The effect of the smoothness prior on the predicted warps in the \textit{ECG200} dataset. Left: no prior. Right: 
    $\lambda_{\sigma}=.01, \lambda_{smooth}=.01$. 
    Color indicates class label. }
    
    \label{fig:prior}
    \end{figure}

\subsection{Diffeomorphisms}\label{cpab}
 As mentioned in~\autoref{Sec:Introduction}, $\Tcal$ needs to be specified. 
In the context of time warping, \emph{diffeomorphisms} is a natural choice~\cite{Mumford:Book:2010:PT}.
  \begin{Definition}
 A ($C^1$) diffeomorphism is a differentiable invertible map with a differentiable inverse. 
  \end{Definition}
 Working with diffeomorphisms usually involves expensive computations.
 In our case, since the proposed method explicitly incorporates them 
in a DL architecture, it is even more important (than in traditional non-DL applications
of diffeomorphisms) to drastically reduce the computational difficulties. 
The reason is that during training, the quantities
$x\mapsto T^\btheta(x)$ and $x\mapsto \nabla_\btheta T^\btheta( x)$
are computed at multiple time points $x$ and for multiple values of $\btheta$. 

As mentioned in~\autoref{Sec:Introduction}, we have chosen to incorporate the CPAB transformation family into DTAN~\cite{Shapira:NIPS:2019:DTAN,Shapira:ICML:2023:RFDTAN}. These warps combine expressiveness and efficiency, making them a natural choice in a DL context~\cite{Hauberg:AISTATS:2016:DA,Skafte:CVPR:2018:DDTN}. 
Other efficient and expressive diffeomorphisms~(\eg, \cite{Zhang:IJCV:2018:fast,Arsigny:BIR:2006,Durrleman:IJCV:2013,Allassonniere:SIAM:2015})
can also be explored in the DTAN context, provided they also offer
an efficient and highly-accurate way to evaluate $x\mapsto\nabla_\btheta T^\btheta( x)$
as CPAB warps do~\cite{Freifeld:TR_CPAB_Derivaitive:2017}.
Below we briefly explain CPAB warps (restricting the discussion to 1D), and refer the reader to~\cite{Freifeld:ICCV:2015:CPAB,Freifeld:PAMI:2017:CPAB,Freifeld:TR_CPAB_Derivaitive:2017}
for more details. 

The name CPAB, short for CPA-Based, is due to the fact that these warps 
are based on Continuous Piecewise-Affine (CPA) velocity fields. 
The term ``piecewise'' is \wrt some partition, denoted by $\Omega$, of the signal's domain into subintervals.
Let $\Vcal$ denote the linear space of CPA velocity fields \wrt such a fixed $\Omega$,
let $d=\dim(\Vcal)$, and let $v^\btheta:\Omega\to\RR$, a velocity field parametrized
by $\btheta\in\Rd$, denote the generic element
of $\Vcal$, where $\btheta$ stands for the coefficient \wrt some basis of $\Vcal$.
The corresponding space of CPAB warps, obtained via integration of elements of $\Vcal$,  is 
\begin{align}
\hspace{-2mm}&\Tcal\triangleq
   \Bigl \{
 T^\btheta:
   x\mapsto \phi^\btheta( x;1)
  \text{ s.t. } \phi^\btheta( x;t) \text{ solves } \nonumber \\ 
\hspace{-2mm} & \phi^\btheta(x;t) = x+\int_{0}^t  v^\btheta(\phi^\btheta(x;\tau))\, 
 \mathrm{d}\tau \text{ where }  v^\btheta\in \Vcal\, 
  \Bigr
 \}\, . \vspace{-5mm}
 \label{Eqn:IntegralEquation}
\end{align}
It can be shown that these warps are indeed ($C^1$) diffeomorphisms~\cite{Freifeld:ICCV:2015:CPAB,Freifeld:PAMI:2017:CPAB}.
See~\autoref{fig:cpab1}
 for a typical warp. While $v^\btheta$
is CPA, $T^\btheta:\Omega\to\Omega$ is not (\eg, $T^\btheta$ is differentiable, unlike $v^\theta$).
CPA velocity fields support an
integration method that is faster \emph{and} more accurate than typical 
velocity-field integration methods~\cite{Freifeld:ICCV:2015:CPAB,Freifeld:PAMI:2017:CPAB}.
The fineness of  $\Omega$ controls the trade-off between expressiveness of $\Tcal$
on the one hand and the computational complexity as well as the dimensionality (\ie, the value of $d=\dim(\btheta)$) on the other 
hand. 

\textbf{Initialization.} Since $\btheta = \bzero$ gives the identity map,
we initialize the final layer of the localization network by sampling the weights from a zero-mean normal
distribution (\ie $\bw \sim \Ncal(\bzero, 10^{-5}))$. 

\textbf{Optional zero-boundary conditions.} If of interest, one can easily restrict 
the CPA fields to vanish at the endpoints of the domain, implying these points will be fixed points
of the resulting warp, \ie $v[0]=v[n]=0$ (see~\citet{Freifeld:PAMI:2017:CPAB} more details). 

\textbf{Optional circularity constraint}. Alternatively, one can enforce a circularity constraint by adding a linear constraint on the CPA velocity field,
making it circularly continuous: $v[0]=v[n]$ (\ie enforcing the velocity at the starting point to be equal to the
velocity at the end-point). See~\cite{kaufman:icip:2021:cyclic} for details.

\textbf{The CPAB Gradient.}
Importantly, in 1D, the \emph{CPAB gradient},
$\nabla_\btheta T^\btheta( x)$, has a closed-form expression which was recently discovered in~\cite{Martinez:ICML:2022:closed}.
The latter is more efficient and stable than the numerical solution and facilitates much faster training and inference time.

\label{sec:cpab}

\begin{figure*}[t]
    \centering
    \def\figwidth{0.17\linewidth }
    \def\trimdim{5mm 5mm 5mm 5mm }
    \def\figdir{figures/barycenters_comparison_panels/}
    \foreach \index in {11,...,20}
    {
        \begin{subfigure}{\figwidth}
            \centering
           {\includegraphics[trim =11mm 6mm 11mm 6mm, clip, width=.98\linewidth]{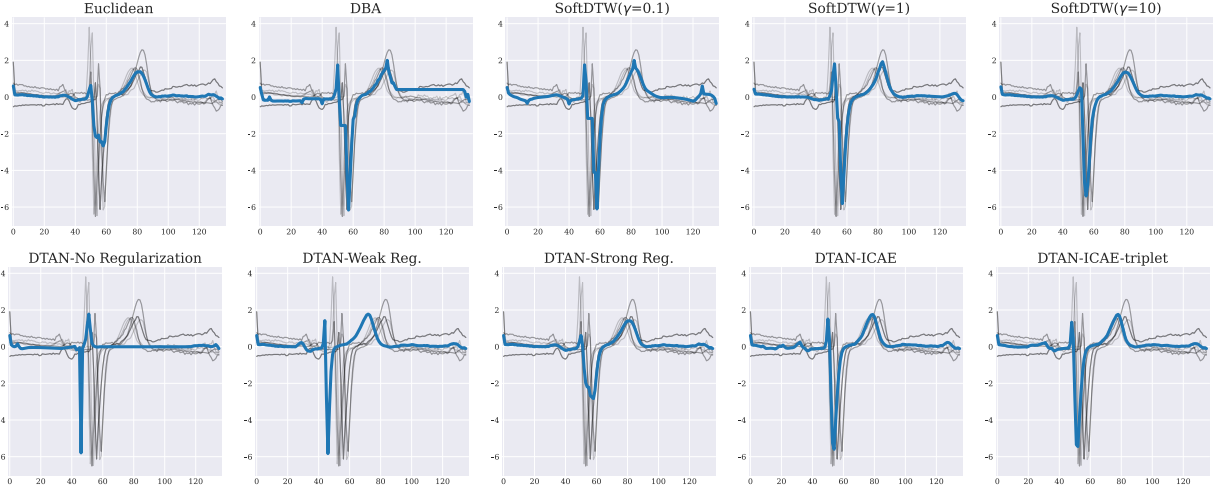}\label{fig:regfree:\index}}

        \end{subfigure}
    }
    \caption{The effect of the regularization HP. The figures shows 10 samples (gray) from the ECGFiveDays dataset with their estimated average (blue), and compares Euclidean averaging, DBA, SoftDTW, and several DTAN methods. DBA requires no HP but falls to poor local minima. SoftDTW's barycenter is severely affected by the choice of its smoothing HP, $\gamma$: $\gamma=0.1$ results in a visible `pinching' effect while $\gamma=10$ smoothens out  peaks/valleys. DBA and SoftDTW are computed per class and do not learn how to generalize to new data, unlike DTAN which is learning-based and requires a single model for all classes. The regularization often used with DTAN has 2 HPs, ($\lambda_{\sigma} ,\lambda_{\mathrm{smooth}}$), where a \emph{weak} regularization ($\lambda_{\sigma} ,\lambda_{\mathrm{smooth}}: .5, .01$) is insufficient and a \emph{strong} regularization ($\lambda_{\sigma},\lambda_{\mathrm{smooth}}:.001, .1$), is too restrictive. $\Lcal_{\mathrm{ICAE}}$ and $\Lcal_{\mathrm{ICAE-triplet}}$ are regularization-free, yet provide barycenters that represent the data well.}
    \label{fig:reg:ecg}
    \end{figure*}

\section{Method}\label{Sec:Method}
To meet our goal, \ie, solving the JA problem while being able to generalize its solution to the alignment of new data, we propose a DL-based method that includes a TTN with diffeomorphic TT layers. Particularly, here we choose $\Tcal$ to be a family of 1D CPAB warps~\cite{Freifeld:ICCV:2015:CPAB,Freifeld:PAMI:2017:CPAB} and incorporate the latter within TT layers. 
Altogether, this lets us propose the first \emph{Diffeomorphic Temporal Alignment Net (DTAN)} for time-series joint alignment.

\subsection{Time Series Joint Alignment}\label{Subsec:Method:objective:function}
\usetikzlibrary{arrows.meta}

\definecolor{lineBlue}{RGB}{57,106,177}
\definecolor{lineOrange}{RGB}{218,124,48}
\definecolor{lineGreen}{RGB}{62,150,81}
\definecolor{lineRed}{RGB}{204,37,41}
\definecolor{lineGray}{RGB}{83,81,84}
\definecolor{linePurple}{RGB}{107,76,154}
\definecolor{lineMaroon}{RGB}{146,36,40}
\definecolor{barBlue}{RGB}{114,147,203}
\definecolor{barOrange}{RGB}{225,151,76}
\definecolor{barGreen}{RGB}{132,186,91}
\definecolor{barRed}{RGB}{211,94,96}
\definecolor{barGray}{RGB}{128,133,133}
\definecolor{barPurple}{RGB}{144,103,167}
\definecolor{barMaroon}{RGB}{171,104,81}

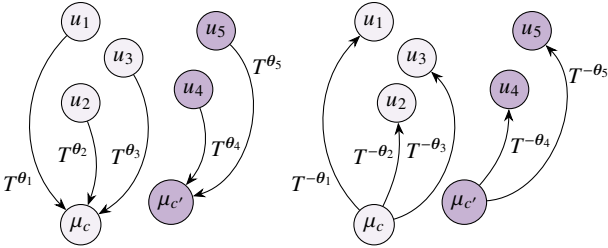
\begin{figure}
    \centering

\begin{tikzpicture}[scale=0.6, every node/.style={transform shape}]

\begin{scope}[every node/.style={circle,draw,minimum size=10,transform shape, font=\Large}]
    \node[fill=barPurple!10] (q2) at (-0.5,5.7) {$u_2$};
    \node[fill=barPurple!10] (q1) at (-0.5,7.5) {$u_1$};
    \node[fill=barPurple!10] (q3) at (0.45,6.7) {$u_3$};
    \node[fill=barPurple!50] (q4) at (2,6) {$u_4$};
    \node[fill=barPurple!50] (q5) at (2.5,7.3) {$u_5$};

    \node[fill=barPurple!10] (A) at (-0.5,3) {$\mu_c$};
    \node[fill=barPurple!50] (B) at (1.5,3.5) {$\mu_{{c'}}$};
\end{scope}

\begin{scope}[>={Stealth[black]},
              every edge/.style={draw=black, transform shape, font=\Large}]
    \path [->] (q1) edge[bend right=45] 
    node[label={$ T^{\btheta_1}$}, shift={(-0.25,-1.75)}] {} (A);
        \path [->] (q2) edge[bend left=20] 
    node[label={$ T^{\btheta_2}$}, shift={(-0.5,-0.2)}] {} (A);
        \path [->] (q3) edge[bend left=45] 
    node[label={$ T^{\btheta_3}$}, shift={(-0.4,-0.5)}] {} (A);
    \path [->] (q4) edge[bend left=30] 
    node[label={$ T^{\btheta_4}$}, shift={(0.5,-0.4)}] {} (B);
    \path [->] (q5) edge[bend left=60] 
    node[label={$ T^{\btheta_5}$}, shift={(0.4,1)}] {} (B);
\end{scope}

\def\offset{6.5}
\begin{scope}[every node/.style={circle,draw,minimum size=10,transform shape, font=\Large}]
    \node[fill=barPurple!10] (U2) at (0+\offset,5.7) {$u_2$};
    \node[fill=barPurple!10] (U1) at (-0.5+\offset,7.5) {$u_1$};
    \node[fill=barPurple!10] (U3) at (0.45+\offset,6.7) {$u_3$};
    \node[fill=barPurple!50] (U4) at (2.5+\offset,6) {$u_4$};
    \node[fill=barPurple!50] (U5) at (3+\offset,7.3) {$u_5$};

    \node[fill=barPurple!10] (A) at (-0.5+\offset,3) {$\mu_c$};
    \node[fill=barPurple!50] (B) at (1.5+\offset,3.5) {$\mu_{{c'}}$};
\end{scope}

\begin{scope}[>={Stealth[black]},
              every edge/.style={draw=black, transform shape, font=\Large}]
    \path [<-] (U1) edge[bend right=45] 
    node[label={$ T^{-\btheta_1}$}, shift={(-0.25,-1.75)}] {} (A);
        \path [<-] (U2) edge[bend left=15] 
    node[label={$ T^{-\btheta_2}$}, shift={(-0.5,-0.2)}] {} (A);
        \path [<-] (U3) edge[bend left=60] 
    node[label={$ T^{-\btheta_3}$}, shift={(-0.5,-0.3)}] {} (A);
    \path [<-] (U4) edge[bend left=20] 
    node[label={$ T^{-\btheta_4}$}, shift={(0.65,-0.3)}] {} (B);
    \path [<-] (U5) edge[bend left=65] 
    node[label={$ T^{-\btheta_5}$}, shift={(0.7,1)}] {} (B);
\end{scope}
\end{tikzpicture}
    \subcaptionbox{Centroids computed using forward warps\label{Fig:Intro:Left}}[0.45\linewidth]{}
    \hfill
    \subcaptionbox{The ICAE loss computed using backward warps\label{Fig:Intro:Right}}[0.45\linewidth]{}
\caption{The Inverse Consistency Averaging Error loss in a two-class example. 
(a) The signals $u_1$, $u_2$, and $u_3$ are in class $c$; $u_4$ and $u_5$ are in class $c'$. Within each class, the centroid
($\mu_c$ or $\mu_{{c'}}$) is obtained by averaging the warped signals ($(u_i\circ T^{\btheta_i})_{i\in\set{1,2,3}}$
or $(u_i\circ T^{\btheta_i})_{i\in\set{4,5}}$)
using the forward warps. (b) The loss is computed using the backward warps; \ie, 
we measure dissimilarity between each $u_i$ and its class centroid, where the latter is first warped backward (``unwarped") using $T^{-\btheta_i}$ (the inverse of $T^{\btheta_i}$). 
}
\label{fig:icae:example}
\end{figure}

Let $u_i$ denote an input signal, let 
$\btheta_i=\btheta_i(u_i,\bw)=
f_{\mathrm{loc}}(u_i,\bw)$
denote the corresponding output of the localization net $f_{\mathrm{loc}}(\cdot,\bw)$
of weights $\bw$,
and let $v_i$, the warped signal, 
denote the result of warping $u_i$ by $T^{\btheta_i}\in\Tcal$;
\ie, $v_i = v_i(u_i,\bw) = u_i \circ T^{\btheta_i}$.
Consider first the case where all the $u_i$'s belong to the same class.
As the variance of the observed $(u_i)_{i=1}^N$ is (at least partially) explained by the latent warps, $(T^{\btheta_i})_{i=1}^N$, 
we seek to minimize the empirical variance of the collection of the warped signals, $(v_i)_{i=1}^N$.
In other words, our data term in this setting is
\begin{align}
\label{eqn:alinment_loss}
 \Lcal_{\mathrm{data}} 
 &\triangleq \tfrac{1}{N} \sum\nolimits_{i=1}^{N} \ellTwoNorm{v_i - \tfrac{1}{N}\sum\nolimits_{j=1}^{N}v_j}^{2} \nonumber \\
 &= \tfrac{1}{N} \sum\nolimits_{i=1}^{N} \ellTwoNorm{u_i \circ T^{\btheta_i} - \tfrac{1}{N}\sum\nolimits_{j=1}^N u_j \circ T^{\btheta_j}}^{2} 
 \nonumber \\  &
 = \tfrac{1}{N} \sum\nolimits_{i=1}^{N} \ellTwoNorm{u_i \circ T^{\btheta_i} - \mu}^{2} \nonumber \\
\end{align}
where  $\ellTwoNorm{\cdot}$ is the $\ell_2$ norm and $\mu$ is the post-alignment average signal.
Note this setting is unsupervised. 

In the multi-class case, $\Lcal_{\mathrm{data}}$ is the sum of the within-class variances, often
called the within-class sum of squares (WCSS):
\begin{align}
\Lcal_{\mathrm{data}} &\triangleq \sum_{k=1}^K 
 \frac{1}{N_k}\sum_{i:y_i=k} \bigg\|u_i \circ T^{\btheta_i} -   \mu_k \bigg\|_{\ell _2}^{2}
\end{align}
where $K$ is the number of classes, $y_i$ takes values in $\set{1,\ldots,K}$
and is the class label associated with $u_i$ (namely: $y_i=j$ if and only if $u_i$ belongs to class $j$), and $N_k=
|\set{i:y_i=j}|$ is the number of examples in class $j$.
In this setting, the learning is partially (or weakly-)supervised: 
the labels, $(y_i)_{i=1}^N$ are known during the learning (but not during the test)
 while the within-class alignment remains unsupervised as in the single-class case. 
The same single network is responsible for aligning 
 each of the classes; \ie, $\bw$ does not vary with $k$. 

Of importance is the fact that, unfortunately, it is possible to reduce $\Lcal_{\mathrm{data}}$ (even to zero!) by severely distorting
the signals such that most of the inter-signal variability concentrates on a small region of the domain and this issue only worsens due to interpolation artifacts.
We now present two complementary methods to avoid this issue:

\textbf{Approach I: Regularizing the predicted warps} by adding: 
 \begin{align}
\label{eqn:loss:reg}
 \Lcal_{\mathrm{reg}}\triangleq  
  \sum\nolimits_{i=1}^N(\btheta_i^T \bSigma_{\mathrm{CPA}}^{-1}\btheta_i)
\end{align}
 where $\bSigma_{\mathrm{CPA}}$ is the CPA covariance matrix
  (proposed by Freifeld \etal~\cite{Freifeld:ICCV:2015:CPAB,Freifeld:PAMI:2017:CPAB}) associated with a
  zero-mean Gaussian smoothness prior over the CPA field.
 Akin to the standard formulation in, \eg, Gaussian processes~\cite{Rasmussen:Book:2004:GP},
$\bSigma_{\mathrm{CPA}}$ has two parameters: $\lambda_{\mathrm{\sigma}}$, which controls the overall variance, and $\lambda_{\mathrm{smooth}}$,
which controls the smoothness of the field. A small $\lambda_{\mathrm{\sigma}}$ favors small warps (\ie, close to the identity) and vice
versa; similarly, 
the larger $\lambda_{\mathrm{smooth}}$ is, the more it favors
CPA velocity fields that are almost purely affine and vice versa, as could be seen in~\autoref{fig:prior}.

The prior also gives another way, an alternative to changing the resolution
of $\Omega$, to control the amount of expressiveness of the warps.
The JA objective function, to be minimized \wrt $\bw$, is
\begin{equation}
  \label{eqn:loss:full}
  \Lcal_{\mathrm{JA}} = \Lcal_{\mathrm{data}} + \Lcal_{\mathrm{reg}}
\end{equation}
which corresponds to~\autoref{eq:JA}, where $D(\cdot)$ is the euclidean distance and $\Rcal(T;\lambda)$ is 
the smoothness prior over the CPA field with HP $\lambda=(\lambda_{\mathrm{\sigma})}, \lambda_{\mathrm{smooth}}$.

However, \emph{optimal regularization is dataset-specific}. For example, penalizing deformations that are too large might
not be ideal in many cases. Likewise, with a temporal smoothness prior, it is hard to determine the ``right" amount of smoothness.  \autoref{fig:reg:ecg} illustrates the critical role of regularization on the  barycenter computation using DBA, SoftDTW, and DTAN.
 Improper values of $\gamma$ (for SoftDTW) or $\lambda_{\sigma},\lambda_{\mathrm{smooth}}$ (for DTAN) usually result in unrealistic warps 
 or overly restrict the warps (\eg, a \emph{strong} prior for DTAN).
This leads us to the second approach, one we have recently introduced in~\cite{Shapira:ICML:2023:RFDTAN}, which is regularization-free.

\textbf{Approach II: Enforcing inverse-consistency} between the post-alignment average sequence and the original samples.
Specifically, we propose a new loss that is minimized when the average sequence is both a minimizer of the variance \emph{and} consistent with its class.  Concretely, we propose the Inverse Consistency Averaging Error loss (ICAE), defined as:
\begin{align}\label{eq:icae}
\Lcal_{\mathrm{ICAE}}&\triangleq \sum_{k=1}^K 
 \tfrac{1}{N_K}\sum_{i:y_i=k}\bigg\| \mu_k\circ T^{-\btheta_i}- u_i\bigg\|\, . 
\end{align}
$\Lcal_{\mathrm{ICAE}}$ measures how well the average signal, $ \mu_k$, fits each signal $u_i$ in its class using the inverse warp $T^{-\btheta_i}$. It does so by first aligning all of the signals in class $k$ using the predicted warps,
then computing their average $ \mu_k$, and finally warping $ \mu_k$ back toward each $u_i$ using $T^{-\btheta_i}$, 
thereby ensuring consistency between them. See~\autoref{fig:icae:example} for an illustration of these two steps.

\begin{figure*}[ht]
\centering
\newcommand{\myWidth}{0.98\linewidth}
\newcommand{\mySubfigWidth}{.19\linewidth}
\newcommand{\mySmallWidth}{1.00\linewidth}
\newcommand{\mySmallSubfigWidth}{.0894\linewidth}
\newcommand{\myL}{0}

\begin{subfigure}{\mySubfigWidth}
  \centering
   \includegraphics[width=\myWidth,trim=5mm 3mm 100mm 23mm,clip]
   {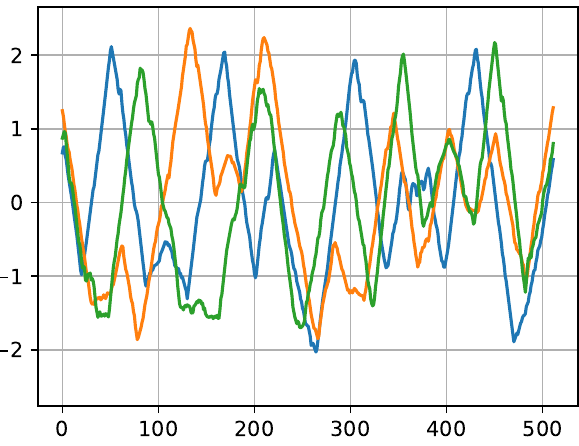}
\end{subfigure}
\foreach \index in {10,20,50,100}
{%
  \begin{subfigure}{\mySubfigWidth}
  \centering
   \includegraphics[width=\myWidth,trim=5mm 3mm 100mm 23mm,clip]
   {figures/training/BeetleFly/no_reg/\index.pdf}
  \end{subfigure}
}
  \begin{subfigure}{\mySubfigWidth}
  \centering
   \includegraphics[width=\myWidth,trim=5mm 3mm 100mm 23mm,clip]
   {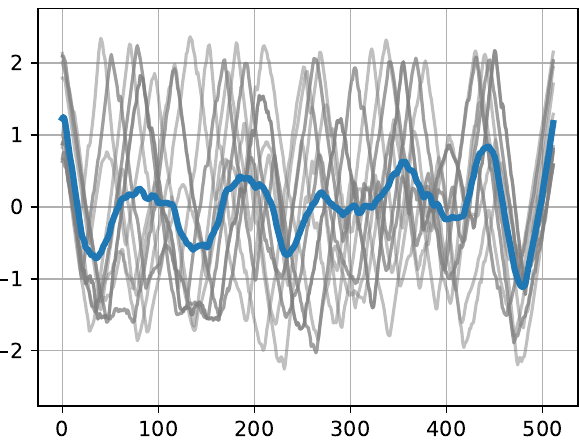}
  \caption{Input data}
  \end{subfigure}
\foreach \index in {10,20,50,100}
{%
  \begin{subfigure}{\mySubfigWidth}
  \centering
   \includegraphics[width=\myWidth,trim=5mm 3mm 100mm 23mm,clip]
   {figures/training/BeetleFly/icae/\index.pdf}
  \caption{Epoch \index}
  \end{subfigure}
}
\caption{Training procedure on the \textit{BeetleFly} dataset. The first column depicts the input data (for better visualization, the top panel shows 3 random signals while the bottom 10 signals and their average are in blue). (\textbf{Top}) The Within-Class Sum of Squares (WCSS) loss reduces variance by applying an unrealistic deformation to the data, resulting in visible `pinching' effect (\ie, bad local minima). (\textbf{Bottom}) The proposed $\mathcal{L}_{\mathrm{ICAE}}$, while requiring no regularization, avoids such an undersired solution by maintaining consistency between the average sequence and its class members.}
\label{fig:training}
\end{figure*}

\emph{A key insight is that ~\autoref{eq:icae} strongly discourages trivial solutions or unrealistic warps as this would result in a poor estimate of $ \mu_k$, which in turn would yield a high discrepancy between it and the original signals.} In other words, the loss favors 
realistic deformations without the need to add a regularization term. This can be seen in~\autoref{fig:training}, where we show the training procedure for the \emph{BeetleFly} dataset for both approaches: WCSS (without regularization) and the ICAE. Minimizing the WCSS distorts the signals to the point that they are no longer recognizable. In contrast, the results under the ICAE retain the key features of the data without enforcing any regularization. 
The full training procedure is described in~\autoref{Alg:training}.
%

%
\subsection{Variable-Length Joint Alignment}\label{Subsec:Method:Variable:Length}
\begin{figure}[t]
\centering
\begin{subfigure}{0.98\linewidth}
 \centering
{\includegraphics[trim = 0mm 8mm 0mm 20mm, clip, width=.95\linewidth]{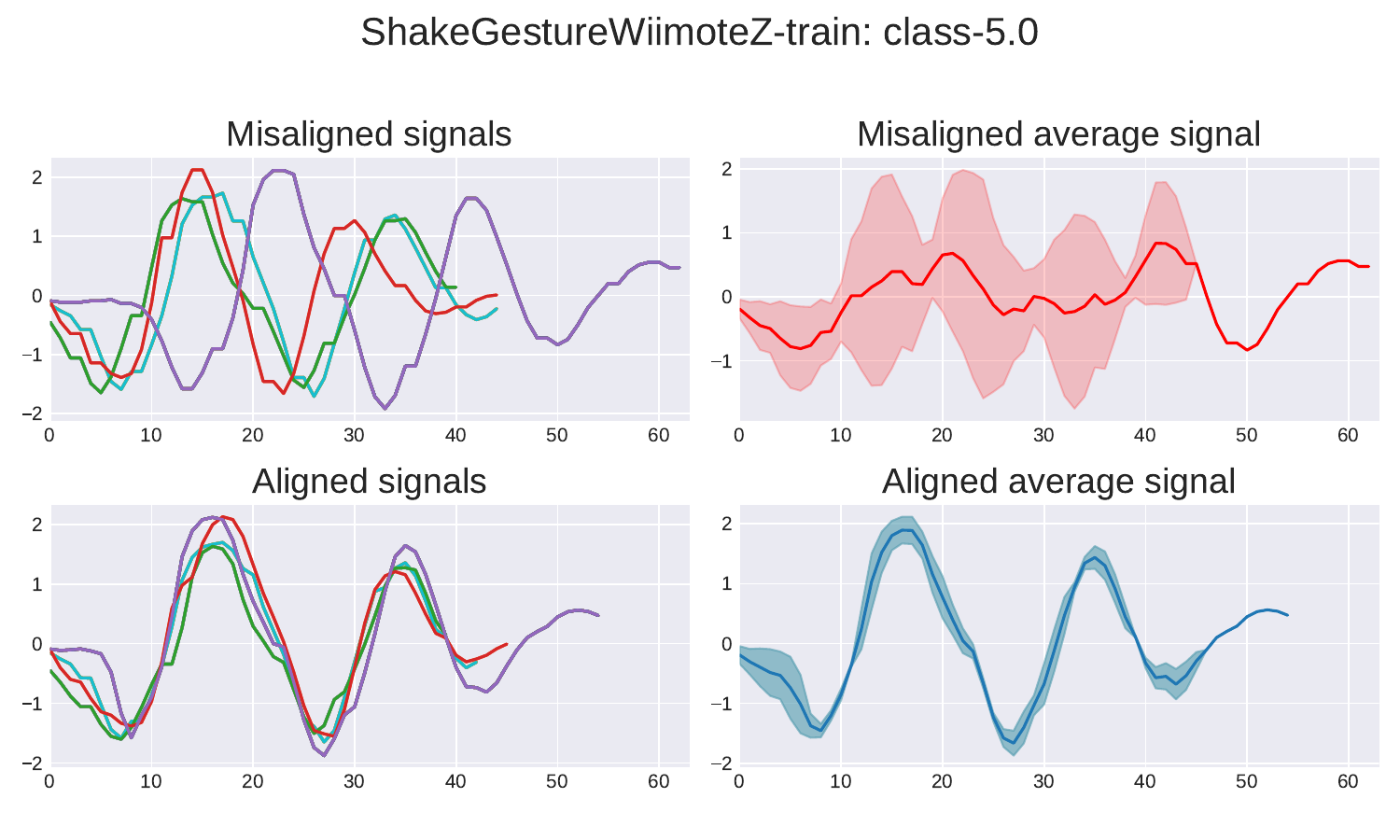}}
\end{subfigure}

\caption{JA of variable-length data (Dataset: \textit{ShakeGestureWiimoteZ}) using the proposed $\mathcal{L}_{\mathrm{ICAE}}$. Shaded area is $\pm$ std.~dev.}
\label{fig:var:len}
\end{figure}

Our proposed $\Lcal_{\mathrm{ICAE}}$ also allows for the JA and averaging of variable-length sequences without having to use a specialized loss function or tweak the boundary conditions on $T^\btheta$ (as mentioned in~\cite{Shapira:NIPS:2019:DTAN, Martinez:ICML:2022:closed} as a hypothetical possibility). Instead, our formulation 
(as well as our code) handles both fixed and variable-length data. It does so in the following manner. First, 
the post-alignment average signal is produced by dividing, at each time step, the sum of the relevant values
by the number of non-missing values.
That is, for each time step $t$ along the duration of the mean signal $\mu$, we compute:
\begin{align}\label{eq:variable:len}
     \mu[t]=\frac{1}{N_{\mathrm{valid}}}\sum_{i:(u_i\circ T^{\btheta_i})[t]\neq \mathrm{null}}^N(u_i\circ T^{\btheta_i})[t]
\end{align}
where $N_{\mathrm{valid}}$ is the number of signals whose domain includes a point mapped to $t$. 
Then, when $ \mu$ is warped backward, ~\autoref{eq:icae} is computed with no modifications. See, \eg,~\autoref{fig:var:len}. 
From an implementation standpoint, we note that any \texttt{null} value in either the input and/or loss would break the computational graph. To avoid \texttt{for-loops} and compute back-propagation in batches, 
it is computationally effective to first pad all samples with zeros (\wrt the longest signal) and create an indicator mask for missing values. The mask is also warped by $T^\btheta$ in~\autoref{eq:variable:len}.

\subsection{Inverse Consistent Centroids Triplet Loss}\label{Sec:Method:Subsec:Triplet}
While $\Lcal_{\mathrm{ICAE}}$ implies consistency, it is agnostic about the separation between different classes.
That said, while metrics such as DTW are completely data-driven, DTAN is learning-based, and can be utilized to learn 
task-driven representations. As such, we introduce the centroid triplet loss into our framework to encourage inter-class separation. Traditionally, \eg in classification tasks, a triplet loss is defined over a triplet ($u^{a}_i, u^p_i, u^n_i$) of an anchor, a positive, and a negative examples, respectively. As our task is intra-class JA and computing class averages (also known as centroids), adopting a centroid-based triplet loss is more adequate here~\cite{doras:2020:prototypical}. We define the \textit{Inverse Consistent Centroids Triplet Loss}  over the triplet ($u^{a}_i,  \mu^p_i,  \mu^n_i$) as 
\begin{align}\label{eq:triplet}
\begin{split}
    &\Lcal_{\mathrm{ICAE-triplet}}(u^{a}_i,  \mu^p,  \mu^n)\triangleq \\ 
    &\max(0, \|u^a_i -  \mu^p\circ T^{- \btheta_i}\|_{\ell _2}^{2} - \|u^a_i -  \mu^n\circ T^{- \btheta_i}\|_{\ell _2}^{2} + \alpha)
\end{split}
\end{align}

where $ \mu^p,  \mu^n$ are \emph{the} positive and \emph{a} negative class centroids, respectively, and $\alpha$ is the margin between them ($\alpha=1$ in all our experiments and is dataset-independent). As both $ \mu^p$ and $ \mu^n$ are compared via an inverse warp, $\Lcal_{\mathrm{ICAE-triplet}}$ does not break the consistency between samples and their mean. The $\Lcal_{\mathrm{ICAE-triplet}}$ is used in tandem with $\Lcal_{\mathrm{ICAE}}$. 


\subsection{Recurrent DTAN}\label{Subsec:Method:RNN}
While often a coarse $\Omega$ 
suffices,
the expressiveness of $\Tcal$ can be increased
using a finer $\Omega$ at the cost of 
computation speed and a higher $d$~\cite{Freifeld:ICCV:2015:CPAB,Freifeld:PAMI:2017:CPAB}. In fact, 
at the limit of
an infinitely-fine $\Omega$, 
any diffeomorphism that is representable by integrating a Lipshitz-continuous stationary velocity field can be approximated by a CPAB diffeomorphism~\cite{Freifeld:ICCV:2015:CPAB,Freifeld:PAMI:2017:CPAB}.
Moreover, CPAB warps do not form a group under the composition operation~\cite{Freifeld:PAMI:2017:CPAB}
(even though they contain the identity warp and are closed under inversion); \ie,
the composition of CPAB warps is a diffeomorphism but usually not CPAB itself.
Thus, a way to increase expressiveness without refining $\Omega$ is by composing  CPAB warps~\cite{Freifeld:PAMI:2017:CPAB}.
Concatenating CPAB warps increases expressiveness beyond $\Tcal$ as it implies a non-stationary velocity field
which is CPA \wrt $\Omega$ and piecewise constant \wrt time.
Compositions increase dimensionality, but the overall cost of evaluating the composed warp scales better (in comparison with refinement of $\Omega$), and it is also easier to infer the $\btheta$'s. 
While this fact was not exploited in~\cite{Skafte:CVPR:2018:DDTN}, we leverage it here as follows.
We propose the Recurrent-DTAN (RDTAN), a net that recurrently applies nonlinear time warps, via diffeomorphic TT layers, to the input signal (\autoref{fig:fig_recurrent}). By sharing the learned parameters by all the TT layers, an RDTAN increases expressiveness without increasing the number of parameters. 
While this is similar to, and inspired by, how Lin \etal~\cite{Lin:CPVR:2017:inverse}
use a recurrent net with affine 2D warps, there is a key difference:
since in the affine case zero-boundary conditions imply degeneracies, they explained they had to propagate warp parameters instead of the warped image as they would have liked. In contrast, as CPAB warps support optional zero-boundary conditions, propagating a warped signal through an RDTAN is a non-issue.

\subsection{Generalization via the Learned Joint-Alignment}\label{Subsec:Method:Generalization}
Once the model is trained, a signal $u$ (regardless whether it is a training or a test signal) is aligned as follows. 
 First set $\btheta=f_{\mathrm{loc}}(u)$; \ie, a forward pass of the net (an operation which is, as is usually the case in DL, simple and very fast). Next, obtain the aligned signal, $v$, by warping $u$ by $T^\btheta$; \ie, set $v=u\circ T^\btheta$. 
 Especially useful and elegant is the fact that, in the multi-class case, the same single net aligns 
 each new test signal, without knowing the label of the latter. This is in sharp contrast
 to other joint-alignment methods (\eg, those based on DBA, SoftDTW, atlases, \etc.)
 that require knowing the label of the to-be-aligned signal.

\subsection{Time Series Averaging}\label{Subsec:Method:Averaging}
The nuisance nonlinear misalignment distorts,
among other things, the sample mean~\cite{wigley:climate:1984:average,Gusfield:Cambridge:1997:Steiner}. As discussed in~\autoref{Sec:previous}, averaging under the DTW distance 
is a commonly-used solution to this issue~\cite{Petitjean:2011:global,Petitjean:2014:dynamic,cuturi:ICML:2014:fast,cuturi:2017:soft}; 
however, such non-learning DTW-based methods are computationally expensive.
This is especially problematic since,
as these methods do not generalize, each batch of new signals requires them to solve another optimization problem (\ie consider the assignment step in the K-means algorithm).
In contrast, as DTAN easily aligns new signals inexpensively and almost instantaneously via a forward pass, it also provides, in the single-class case, a mechanism 
for quickly averaging a new collection of previously-unseen signals.
In other words, this is nothing more than computing the sample mean
of the warped test data: 
\begin{align}
\label{eqn:average}
\mu = \tfrac{1}{N}\sum\nolimits_{i=1}^{N}v_i = \tfrac{1}{N}\sum\nolimits_{i=1}^N u_i \circ T^{\btheta_i}\, .
\end{align}
\newcommand{\pluseq}{\mathrel{{+}{=}}}

{    \SetKwComment{Comment}{}{}
\SetInd{4mm}{0.5mm}

\begin{algorithm}[t]
    \KwIn{$N_{\mathrm{epochs}}$, $f_{\mathrm{loc}}$}
    
    \KwData{$(u_i,y_i)_{i=1}^N$}
    
    \KwOut{$f_{\mathrm{loc}}(\cdot)$, trained for joint alignment}
    

    \For{each epoch and  each batch $j\in \set{1,\ldots,N_\mathrm{batches}}$}  { 
        $\Lcal_{\mathrm{batch}} \gets 0$

        $(u_i, y_i)_{i=1}^{N_j} \gets \text{batch}_{j}$

       $(\btheta_i)_{i=1}^{N_j} \gets (f_{\mathrm{loc}}(u_i))_{i=1}^{N_j}$

        \For{$k\in \set{1,\ldots,K}$}{
        $ \mu_k=\frac{1}{N_k}\sum_{i:y_i=k}(u_i\circ T^{\btheta_i})$ 
        
        $\Lcal_{\mathrm{ICAE}}=
        \frac{1}{N_K}\sum_{i:y_i=k}\| \mu_k\circ T^{-\btheta_i}-u_i\|_{\ell _2}^{2}$
        
        $\Lcal_{\mathrm{batch}} \pluseq \Lcal_{\mathrm{ICAE}}$
        }



    Perform an optimization step to minimize $\Lcal_\mathrm{batch}$
    }

    \caption{The JA training  with an ICAE loss}   \label{Alg:training}
\end{algorithm}
}

\subsection{Multi-task learning}\label{Subsec:Method:Multitask}
In addition to the aforementioned objective functions, this work introduces DTAN to the notion of multitask learning. 
Inspired by the recent success of multitask learning in the context of time-series averaging~\cite{terefe:ICTAI:2020:time}, 
we propose to incorporate a classification objective as a second task in the DTAN framework.
As stated in~\cite{terefe:ICTAI:2020:time}, the classification objective is set to mitigate the chances 
of overlapping means between classes and serves as a complementary approach to $\Lcal_{\mathrm{ICAE-triplet}}$. Thus, to increase separability between classes, we propose to add a
cross-entropy term to \autoref{eqn:loss:full}:
\begin{equation}
    \label{eqn:ce}
    \Lcal_{\mathrm{ce}} \triangleq -\sum_{i=1}^{N} y_i \log \tilde{y}_i \, .
\end{equation}
where $y_i$ are the true class labels and $\tilde{y}_i$ are the predicted ones.
In terms of architecture, we attach a fully-connected layer with a SoftMax activation to the penultimate layer (\ie the embedding) of $f_{\mathrm{loc}}(\cdot)$.
Given a penultimate layer of $dim=M$, the additional classification head only adds $M\times K$ parameters to the final model. 
The classification framework is supervised \wrt the class labels, 
but still unsupervised \wrt the time-series alignment.

To control the trade-off between joint alignment and classification/separability, we introduce a hyperparameter $\lambda_{ce}$,
which is set to 1 by default. Thus, the multitask loss function is defined as:

\begin{align}
    \Lcal_{\mathrm{multi}} \triangleq 
    \Lcal_{data} + \Lcal_{reg}+ \lambda_{ce} \Lcal_{\mathrm{ce}}
\end{align}

In the case of RDTAN (to be discussed in \autoref{Subsec:Method:RNN}), the classification head is 
used only at the last recurrence of RDTAN.

\subsection{Implementation}\label{Subsec:Method:Implementation}
We adapted, to the 1D case, the implementation from \texttt{libcpab}~\cite{Detlefsen:2018:libcpab}
of the CPAB transformer layer, CPAB gradient, and the PyTorch C++ API. The close-formed gradient proposed in~\cite{Martinez:ICML:2022:closed} is given by the \texttt{DIFW} package.
We have implemented the CPAB regularization term, objective functions, and DTAN variants in PyTorch.
 Our code is available at 
\href{https://github.com/BGU-CS-VIL/RF-DTAN}{
 \texttt{https://github.com/BGU-CS-VIL/RF-DTAN}}.

\begin{figure*}[t]
\centering
\newcommand{\myWidth}{0.98\linewidth}
\newcommand{\mySubfigWidth}{.32\linewidth}
\newcommand{\mySmallWidth}{1.00\linewidth}
\newcommand{\mySmallSubfigWidth}{.0894\linewidth}
\newcommand{\myL}{0}
\foreach \index in {0,1,2}
{%
  \begin{subfigure}{\mySubfigWidth}
  \centering
   \includegraphics[width=\myWidth,trim=5mm 5mm 5mm 22mm,clip]
   {figures/alignment/CBF/CBF-train\index.pdf}
  \caption{Class \index}
  \end{subfigure}
}
\label{fig:CBF:train}
\end{figure*}
%
%
\begin{figure*}[t]
\centering
\newcommand{\myWidth}{0.98\linewidth}
\newcommand{\mySubfigWidth}{.48\linewidth}
\newcommand{\mySmallWidth}{1.00\linewidth}
\newcommand{\mySmallSubfigWidth}{.0894\linewidth}
\newcommand{\myL}{0}
\foreach \index in {0,1}
{%
  \begin{subfigure}{\mySubfigWidth}
  \centering
   \includegraphics[width=\myWidth,trim=5mm 5mm 5mm 22mm,clip]
   {figures/alignment/ECG200/ECG200-train\index.pdf}
  \caption{Class \index}
  \end{subfigure}
}
\label{fig:ECG200:train}
\end{figure*}
%
%
\begin{figure*}[t]
\centering
\def\figwidth{0.98\textwidth}
\newcommand{\myWidth}{0.98\linewidth}
\newcommand{\mySubfigWidth}{.32\linewidth}
\newcommand{\mySmallWidth}{1.00\linewidth}
\newcommand{\mySmallSubfigWidth}{.0894\linewidth}
\newcommand{\myL}{0}
\foreach \index in {0,1,2}
{%
  \begin{subfigure}{\mySubfigWidth}
  \centering
   \includegraphics[width=\myWidth,trim=5mm 5mm 5mm 22mm,clip]
   {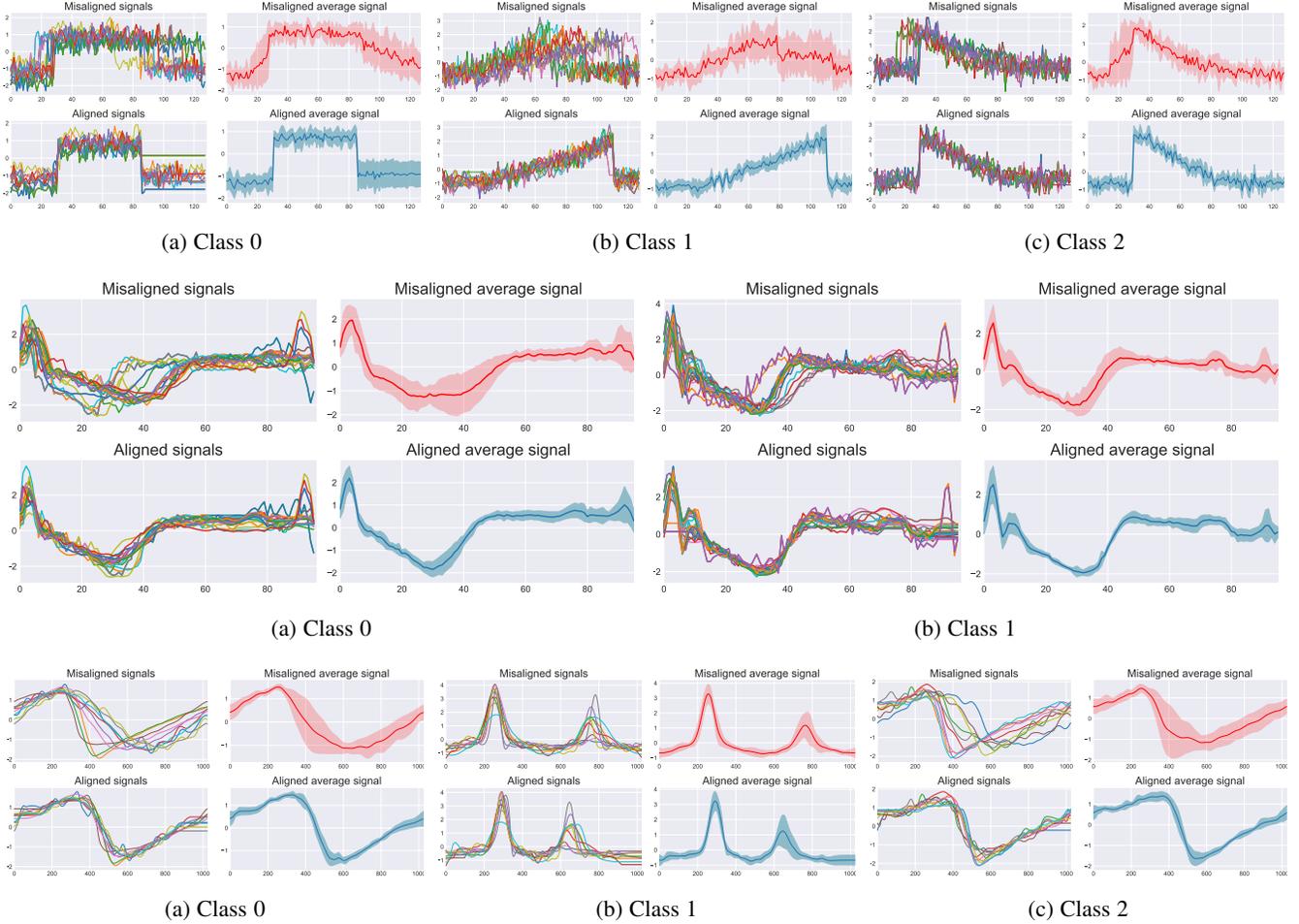}
  \caption{Class \index}
  \end{subfigure}
}
\caption{Joint alignment and averaging of the (top) \textit{CBF}, (middle) \textit{ECG200}, and (bottom) \textit{StarLightCurves} datasets using $\mathcal{L}_{\textrm{ICAE}}$. The shaded area corresponds to $\pm\sigma$.}
\label{fig:JA:results}
\end{figure*}

\section{Experiments and Results}\label{Sec:Results}
The evaluation of our approach was conducted on both synthetic and real-world data, using the popular \emph{UCR time-series classification archive} benchmark which contains 128 datasets.~\autoref{fig:JA:results} depicts JA results on some of the UCR datasets.
The rest of the section is structured as follows.
First, we evaluate the effect of predicting non-stationary velocity fields via RDTAN in~\autoref{Sec:Results:sub:RDTAN}. 
In~\autoref{Sec:Results:sub:NCC} we compare DTAN to state-of-the-art time-series JA and averaging methods on the UCR archive in a series of experiments. 
In~\autoref{Sec:Results:sub:timming} we show the computational benefits of using DTAN compared with DTW-based approaches. 
In~\autoref{Sec:Results:sub:MTDTAN} we provide new details regarding the effect of $f_{\mathrm{loc}}(\cdot)$ and multi-task learning on JA.
Finally,~\autoref{Sec:Results:sub:PCA} details how DTAN improves PCA.


\begin{figure*}[t]
    \centering
    \def\figwidth{0.92\linewidth}
    
    \begin{tikzpicture}
        \node[anchor=south west, inner sep=0] (image) at (0,0) {
            \includegraphics[trim = 0mm 2mm 9mm 2mm, clip, width=\figwidth]{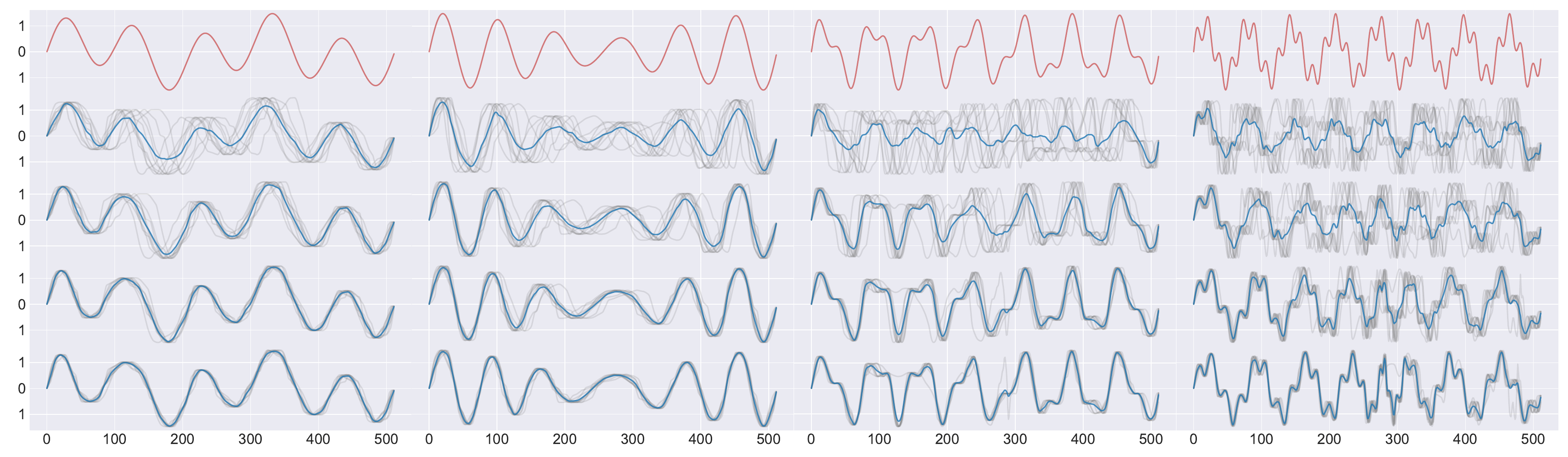}
        };
        
        \foreach \row/\label in {1/(e), 2/(d), 3/(c), 4/(b), 5/(a)} {
            \node[rotate=0, anchor=south, font=\scriptsize] at ([xshift=-1mm, yshift=(\row-3)*10mm]image.west) {\label};
        }
    \end{tikzpicture}
    
    \caption{Recurrent DTAN (RDTAN) JA of synthetic data. (a) latent source and (b) 10 perturbed signals (gray) and their average (blue). (c)-(e) RDTAN output at each recurrence, where the latent source signals are gradually recovered by finding the perturbed signals JA.}
    \label{fig:fig_recurrent}
\end{figure*}

\subsection{Recurrent DTANs}\label{Sec:Results:sub:RDTAN}

\autoref{fig:fig_recurrent} displays the JA of synthetic data using RDTAN.
We generated the synthetic data by perturbing four synthetic signals with random warps obtained via cumulative distribution functions sampled from a Dirichlet-distribution prior, as detailed in~\cite{Shapira:NIPS:2019:DTAN}. 
The top row presents the original latent sequences in red, the second row the perturbed synthetic signals in gray and their average in blue, and each of the subsequent three rows illustrates the alignment achieved in successive iterations of RDTAN. All four classes (columns) are aligned using the same single model. Consistent with our earlier discussion, the latent average and the latent warps are unknown during both training and testing, yet DTAN successfully recovers the latent signals via the successful JA. 

\begin{figure}[t]
    \centering
    \def\figwidth{0.9\linewidth } 
    \begin{subfigure}{\figwidth}
     \centering
    {\includegraphics[trim = 1mm 0mm 108mm 15mm, clip, width=0.7\linewidth]{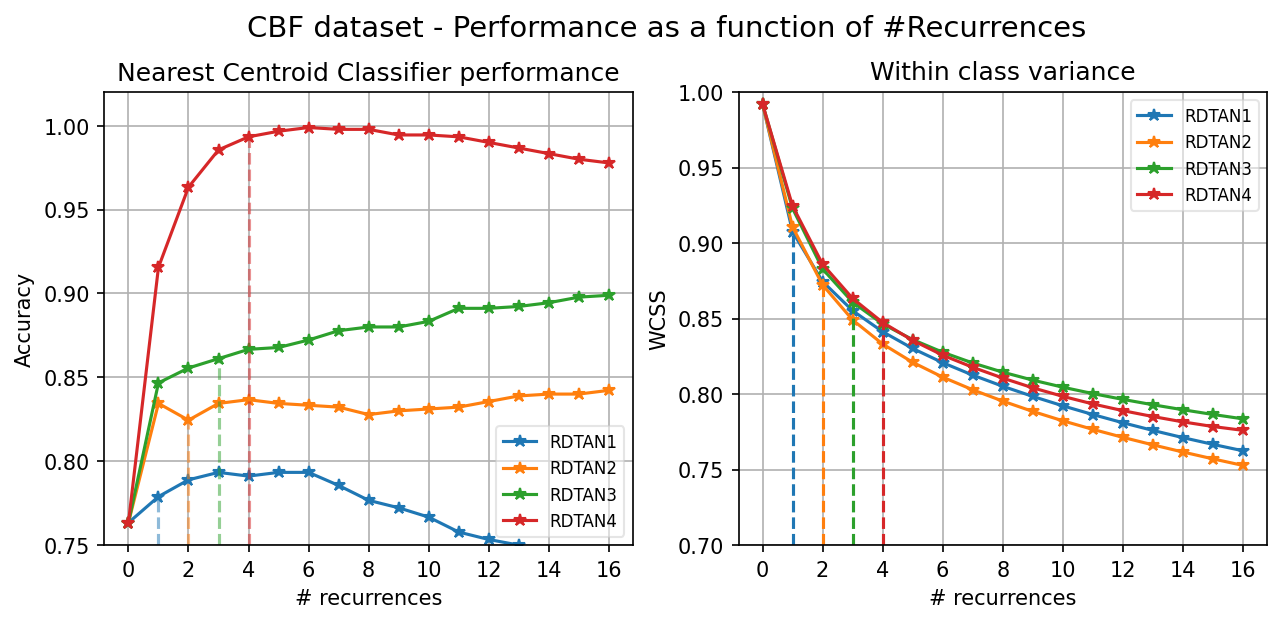}}
    \end{subfigure}

    \caption{Nearest Centroid Classifier (NCC) performance as a function of the number of recurrent wraps applied by RDTAN during inference, evaluated on the \textit{CBF} dataset. 
    Dashed vertical lines indicate the number of recurrences used during training.}
    \label{fig:rnn:compare}
    \end{figure}

RDTAN's performance was assessed using the same recurrence number as in training. In certain cases, RDTAN benefits from applying extra warps beyond the training count. This improvement is feasible because the learned parameters are shared across all warps, akin to standard RNNs.~\autoref{fig:rnn:compare} demonstrates the impact of increasing the number of applied warps (up to 16) on the CBF dataset (which is a part of the UCR archive~\cite{Dau:2019:ucr}), particularly in terms of NCC accuracy (defined below). The findings reveal that (a) DTAN, without recurrences, struggles with additional ones as it lacks relevant training, and (b) RDTAN either maintains robustness (RDTAN2) or shows enhanced performance (RDTAN3/4) with an increased number of recurrences.

\subsection{Nearest Centroid Classification (NCC)}\label{Sec:Results:sub:NCC}
The most updated version~\cite{Dau:2019:ucr} of the UCR archive has 128  datasets with inter-dataset variability in the number of samples, signal length, application domain, and the number of classes. Eleven of those datasets also present intra-dataset variability of the signal length; such datasets are referred to as variable-length (VL) datasets.
In all of the experiments, we used the train/test splits provided by the archive. 
To quantify performances we used, as is customary, 
the NCC accuracy. This performance index is viewed as an evaluation metric for measuring how well each centroid describes its class members (and thus, implicitly, also measures the JA quality). 
The NCC framework has 2 steps: 1) compute the centroid, $\mu_k$, for each class $k$ of the \emph{train} set; 2) label each \emph{test} sample by the class of its closest centroid. As we explain below,~\autoref{table:ncc}, which summarizes the NCC results, is divided into several parts. The full results, together with many illustrative figures, train/test comparison, and additional evaluations, appear in our Supplemental Material \textbf{(SupMat)}.

In all of our DTAN experiments, training was done via the Adam optimizer~\cite{Kingma:arxiv:2014:Adam} for 1500 epochs, batch size of 64, $N_p$ (the number of subintervals in the partition of $\Omega$) was 16, and the scaling-and-squaring parameter (used by \texttt{DIFW}) was 8. These values were previously reported to yield the highest number of \emph{Wins} in~\cite{Martinez:ICML:2022:closed}. 
While in~\cite{Shapira:NIPS:2019:DTAN} we have used RDTAN, in~\citet{Martinez:ICML:2022:closed} the authors stacked TCNs sequentially. In this study, we fixed the number of recurrences to 4 as we did not find it necessary to stack InceptionTime models.
The PyTorch \texttt{TSAI} implementation of the InceptionTime was taken from~\cite{Ignacio:tsai}.
For DTW, DBA, and SoftDTW we used the \texttt{tslearn} package~\cite{tavenard:2017:tslearn}. 
\subsubsection{Part 1: 84 datasets -- allowing an extensive HP search (previously-reported results).}
\begin{table*}[t]
\caption{Nearest Centroid Classification Accuracy.}
\label{table:ncc}
\vskip 0.15in
\begin{center}
\begin{footnotesize}
\begin{sc}
\begin{tabular}{lcccccccr}
\toprule
Method & Objective & NCC$_{\mathrm{median}}$ & NCC$_{\mathrm{best}}$ & $\#$configs & $\#$Datasets & $\#$experiments \\
\midrule
\multicolumn{7}{c}{Part 1: Allowing HP Search (previously-reported results)} \\
\midrule
Euclidean            & \textit{N/A} & - & 0.611 & 1   & 84 & 84  \\
DBA~\cite{Petitjean:2011:global}                  & DTW          & - &0.657 & 1   & 84 & 84  \\
SoftDBA~\cite{cuturi:2017:soft}              & SoftDTW & - & 0.703 & 9   & 84 & 756 \\
SoftDBA~\cite{Blondel:2021:differentiable}          & SoftDTW-div & - & 0.708 & 9   & 84 & 756   \\
DTAN$_{\mathrm{libcpab}}$~\cite{Shapira:NIPS:2019:DTAN}     & WCSS + Reg & - & 0.705 & 12  & 84 & 1008 \\
ResNet-TW~\cite{huang:2021:residual}            & WCSS + Reg & - & 0.711 & 20  & 84 & 1680 \\
DTAN$_{\mathrm{DIFW}}$~\cite{Martinez:ICML:2022:closed}        & WCSS + Reg & - & \textbf{0.749} & 96  & 84 & 8064 \\
\midrule
\multicolumn{7}{c}{Part 2: Single HP Configuration in all datasets (same UCR datasets as reported by other works above)} \\ 
\midrule
DTAN$_{\mathrm{DIFW}}$~\cite{Shapira:NIPS:2019:DTAN, Martinez:ICML:2022:closed} & WCSS + Reg  & 0.604 & 0.607	 & 1   & 84 & 84 \\
DTAN$_{\mathrm{DIFW}}$~\cite{Shapira:ICML:2023:RFDTAN} &$\Lcal_{\mathrm{ICAE}}$  & 0.665 & 0.694 & 1   & 84 & 84 \\
DTAN$_{\mathrm{DIFW}}$~\cite{Shapira:ICML:2023:RFDTAN} & $\Lcal_{\mathrm{ICAE-triplet}}$ &  \textbf{0.707} & \textbf{0.739}  & 1 & 84 & 84  \\
\midrule
\multicolumn{7}{c}{Part 3: Single HP Configuration in all datasets (including additional newer fixed-length UCR datasets)} \\ 
\midrule
DTAN$_{\mathrm{DIFW}}$~\cite{Shapira:NIPS:2019:DTAN, Martinez:ICML:2022:closed} & WCSS & 0.609 & 0.65 & 1  & 117 & 117  \\
DTAN$_{\mathrm{DIFW}}$~\cite{Shapira:NIPS:2019:DTAN, Martinez:ICML:2022:closed} & WCSS + Reg & 0.603 & 0.605 & 1  & 117 & 117  \\
DTAN$_{\mathrm{DIFW}}$~\cite{Shapira:ICML:2023:RFDTAN}  & $\Lcal_{\mathrm{ICAE}}$ & 0.656 & 0.686 & 1   & 117 &117 \\
DTAN$_{\mathrm{DIFW}}$~\cite{Shapira:ICML:2023:RFDTAN}  &$\Lcal_{\mathrm{ICAE-triplet}}$ & \textbf{0.709} & \textbf{0.741}  & 1   & 117 & 117  \\

\midrule
\multicolumn{7}{c}{Part 4: Single HP Configuration in all datasets (full updated UCR archive, including variable-length datasets)} \\ 
\midrule
DTAN$_{\mathrm{DIFW}}$~\cite{Shapira:ICML:2023:RFDTAN}  & $\Lcal_{\mathrm{ICAE}}$ & 0.623 &0.653 & 1   & 128 & 128 \\
DTAN$_{\mathrm{DIFW}}$~\cite{Shapira:ICML:2023:RFDTAN}  & $\Lcal_{\mathrm{ICAE-triplet}}$ & \textbf{0.67} & \textbf{0.701}  & 1   & 128 & 128  \\
\midrule
\bottomrule
\end{tabular}
\end{sc}
\end{footnotesize}
\end{center}
\vskip -0.1in
\end{table*}
An older version~\cite{Chen:UCR:Archive:2015} of the UCR archive had only 85 datasets (a subset of the 128 mentioned above). Several previous works reported results on only 84 datasets out of those 85, possibly due to the size of the largest dataset.
Part 1 of~\autoref{table:ncc} contains the results, on those 84 datasets, obtained by several key methods, as reported by their authors, as well as those obtained by a simple Euclidean averaging (\ie, a no-alignment baseline). The methods are DBA, SoftDTW, DTAN$_{\mathrm{libcpab}}$, ResNet-TW, and  DTAN$_{\mathrm{DIFW}}$. 
The regularization-free DBA requires no HP configurations. 
The SoftDTW methods have one HP for controlling the smoothness. Their results, reported in~\cite{Blondel:2021:differentiable}, were obtained by those authors using cross-validation. 
The other works~\cite{Shapira:NIPS:2019:DTAN,huang:2021:residual,Martinez:ICML:2022:closed}
reported only their best results
across different configurations. 
In~\cite{Shapira:NIPS:2019:DTAN} we have evaluated  DTAN$_\mathrm{libcpab}$ using 12 different configurations per dataset
(4 configurations for $(\lambda_{\sigma}, \lambda_{\mathrm{smooth}})$ and 3 different numbers of recurrences). In~\cite{huang:2021:residual}, ResNet-TW used the same regularization configurations
as in~\cite{Shapira:NIPS:2019:DTAN}, but also tested varying numbers of ResNet blocks (4 to 8) per dataset.
\citet{Martinez:ICML:2022:closed} evaluated  DTAN$_{\mathrm{DIFW}}$ using 96 different configurations 
(various options of $\lambda_{\sigma}, \lambda_{\mathrm{smooth}}, N_p, \#$stacked TCNs, boundary conditions, and the scaling-and-squaring parameter) per dataset. 
We note that: 1) tuning $N_p$ and the boundary conditions is another form of tweaking
the regularization; 2) as stated in (the supplemental material of)~\cite{Martinez:ICML:2022:closed}, their reported results were chosen among those 96 configurations, per dataset, based on the best performance on the test set.

\subsubsection{Part 2: Regularization vs regularization-free DTAN}
Part 1 of~\autoref{table:ncc} suggests that increasing the number of tried HP configurations translates to better performance due to the large variability across the UCR datasets. 
However, the compact summary in Part 1 of~\autoref{table:ncc} also hides an inconvenient truth:
there is no \emph{one-size-fits-all} configuration. For example, ~\citet{Martinez:ICML:2022:closed} produced the best performance but this is largely due to an expensive search over a large number of HP configurations.
In fact, inspecting the full results of either DTAN$_{\mathrm{libcpab}}$, ResNet-TW, or DTAN$_{\mathrm{DIFW}}$, reveals that 
the optimal choice of HP varies across the datasets and affects results drastically.  

To demonstrate this crucial point, we ran a new set of experiments. 
We picked the HP configuration 
that according to~\cite{Martinez:ICML:2022:closed}
achieved the highest number of wins among their 96 configurations.  
Next, using that configuration we ran,  on those 84 datasets, exactly the same DTAN but with 3 different losses:
1) WCSS plus the smoothness regularization ($\lambda_{\sigma}$ and $\lambda_{\mathrm{smooth}}$, 0.001 and 0.1, respectively);
2) our proposed $\Lcal_{\mathrm{ICAE}}$;
3) our proposed $\Lcal_{\mathrm{ICAE-triplet}}$.
In the last 2 cases, which are regularization-free, 
the values of $\lambda_{\sigma}$ and $\lambda_{\mathrm{smooth}}$ from that configuration were ignored. 
In all 3 cases, we used DTAN$_{\mathrm{DIFW}}$
with the same InceptionTime backbone~\cite{Ignacio:tsai} (in all 3 cases this gave better results than using a TCN). 
To account for random initializations and the stochastic nature of 
DL training,
in each of the 3 cases we performed 5 runs on each dataset and report both the median and best results;
see part 2 in~\autoref{table:ncc}.
The results illustrate the merits of our regularization-free approach:
a single HP configuration for the regularization, even the one stated as the best, does not properly fit the entirety of the UCR datasets. 
In contrast, dropping the regularization term and using our $\Lcal_{\mathrm{ICAE}}$ increases performance by a large margin, which is only further increased when utilizing $\Lcal_{\mathrm{ICAE-triplet}}$, 
which increases separability between class centroids (a feat current DTW-based methods are incapable of) and achieves SOTA results. 

\subsubsection{Part 3 \& 4: Using a single HP configuration in all of the 128 datasets.}
To produce the results in part 3 of~\autoref{table:ncc}, we again repeated the procedure from part 2, 
except that 1) we added another case where the loss is only WCSS with no regularization,
and 2) the results, on 117 datasets, also take into account additional fixed-length datasets that were added
in the newer UCR archive. 
The results in, and conclusions from,  Part 3 are consistent with Part 2. WCSS did slightly better than WCSS+Reg, probably since even though it distorts the signals, it makes it a bit easier (than in the WCSS+Reg case) to differentiate between classes. In any case, our losses outperform both of these methods. \textbf{Part 4} extends the results of Part 3 by adding, for the DTANs with our proposed losses, the 11 VL datasets (for a total of 128). 

\begin{table*}[t]
\centering
\caption{Timing comparison. (Top) During the fitting/training step, SoftDTW/DBA are computed per class while DTAN$_{\mathrm{ICAE}}$ uses one model for all classes. (Middle) During inference, 30 new samples are averaged. Soft/DBA needs to be called again as it is optimization-based, while DTAN$_{\mathrm{ICAE}}$ requires a single forward pass. (Bottom) Finally, each new sample is compared to its train-set barycenter using the corresponding metric. \textbf{N/A = Out Of Memory} (on a machine with 12 CPU cores and 32Gb RAM).} 
\label{tab:timing}
\footnotesize
\begin{tabular}{lllllllll}
\toprule
Dataset         & $N_{samples}$ & $N_{class}$ & Length & DBA             & SoftDTW$_{\gamma=0.01}$     & SoftDTW$_{\gamma=0.1}$   & SoftDTW$_{\gamma=1}$         & $\mathrm{DTAN_{ICAE}}$
\\ \midrule
\multicolumn{9}{c}{Training time - full train set  (sec)}                                                                                              \\ \midrule
ECGFiveDays     & 23            & 2           & 136    & 0.90            & 0.65            & 0.64             & \textbf{0.31}           & 157.39          \\
Yoga            & 300           & 2           & 426    & 52.4104         & 265.493         & 283.566          & \textbf{50.3923}        & 565.65          \\
StarLightCurves & 300           & 3           & 1024   & 1140.79         & 3399.90         & 964.21           & \textbf{441.33}         & 2657.20         \\
HandOutlines    & 1000          & 2           & 2709   & N/A             & N/A             & N/A                 & N/A          & \textbf{6483.50}         \\ \midrule
\multicolumn{9}{c}{Inference time, averaged over 5 runs (sec)}                                                                       \\\midrule 
ECGFiveDays     & 30            & 1           & 136    & 0.3$\pm$0.03    & 3.39$\pm$1.32   & 2.22$\pm$0.36    & 0.73$\pm$0.25  & \textbf{0.02$\pm$0.013} \\
Yoga            & 30            & 1           & 426    & 4.21$\pm$0.9    & 31.46$\pm$5.92  & 27.58$\pm$4.33   & 4.73$\pm$0.38  & \textbf{0.02$\pm$0.01}   \\
StarLightCurves & 30            & 1           & 1024   & 19.08$\pm$3.06  & 80.52$\pm$12.71 & 61.5$\pm$22.07   & 15.2$\pm$0.15  & \textbf{0.02$\pm$0.01}   \\
HandOutlines    & 30            & 1           & 2709   & 70.69$\pm$28.39 & 209.2$\pm$58.93 & 155.53$\pm$18.37 & 68.54$\pm$0.3  & \textbf{0.04$\pm$0.02}   \\ \midrule
\multicolumn{9}{c}{Distance to barycenter using the corresponding metric, averaged over 5 runs (sec)}                                                       \\ \midrule 
ECGFiveDays     & 30            & 1           & 136    & 0.05$\pm$0.0    & 0.04$\pm$0.0    & 0.04$\pm$0.0     & 0.05$\pm$0.0   & \textbf{0.024$\pm$0.0}   \\
Yoga            & 30            & 1           & 426    & 0.09$\pm$0.0    & 0.25$\pm$0.0    & 0.28$\pm$0.0     & 0.3$\pm$0.0    & \textbf{0.024$\pm$0.0}   \\
StarLightCurves & 30            & 1           & 1024   & 0.11$\pm$0.01   & 1.39$\pm$0.01   & 1.61$\pm$0.0     & 1.76$\pm$0.0   & \textbf{0.023$\pm$0.0}   \\
HandOutlines    & 30            & 1           & 2709   & 0.4$\pm$0.01    & 10.03$\pm$0.01  & 11.5$\pm$0.03    & 12.54$\pm$0.07 & \textbf{0.045$\pm$0.002} \\
\bottomrule
\end{tabular}%
\end{table*}

\subsection{Computation-time Comparison}\label{Sec:Results:sub:timming}
A key advantage of learning-based approaches is fast inference on new data. 
We performed several timing comparisons
between DBA, SoftDTW (whose HP, $\gamma\in\{0.01,0.1,1\}$, must be searched in each dataset), and DTAN, trained with the proposed $\Lcal_{\mathrm{ICAE}}$. We used a machine with 12 CPU cores, 32Gb RAM, and an RTX 3090 GPU card. We chose a subset of the UCR archive, spanning different lengths and sample sizes, and compared the time it took to compute the centroids on the entire train set. Since DBA and SoftDTW are optimization-based we provide timing for two approaches: (1) barycenter computation time of a new batch ($N=30$, average of 5 runs) and (2) computing DTW/SoftDTW between the batch and its barycenter (which, after warping, can be averaged again). For DTAN, this is just the inference time. \autoref{tab:timing} presents the result. On training data, for small datasets (in terms of $n, N$), SoftDTW/DBA is faster than DTAN, but this trend is reversed for the large ones. \textbf{SoftDTW and DBA run out of memory} on the largest dataset (\texttt{HandOutlines}). During inference, using DTAN is \emph{orders of magnitude faster} (x$10$--x$10^{4}$) than recomputing barycenters, and, on the larger datasets, is x$10$ faster than computing DTW/SoftDTW. 

\subsection{Multi-task Learning and Backbone Comparison}~\label{Sec:Results:sub:MTDTAN}
\pgfplotsset{compat=1.11,
    /pgfplots/ybar legend/.style={
    /pgfplots/legend image code/.code={%
       \draw[##1,/tikz/.cd,yshift=-0.25em]
        (0cm,0cm) rectangle (3pt,0.8em);},
   },
}

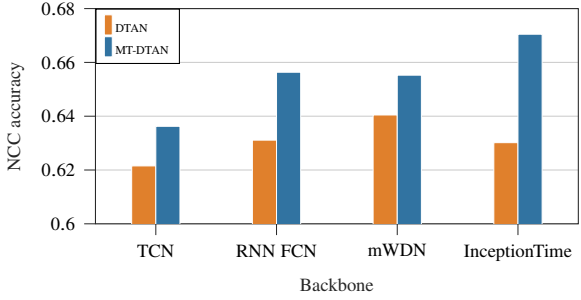
\begin{figure}
\begin{tikzpicture}

\definecolor{brown1926061}{RGB}{192,60,61}
\definecolor{darkslategray38}{RGB}{38,38,38}
\definecolor{lightgray204}{RGB}{204,204,204}
\definecolor{mediumpurple147113178}{RGB}{147,113,178}
\definecolor{peru22412844}{RGB}{224,128,44}
\definecolor{seagreen5814558}{RGB}{58,145,58}
\definecolor{steelblue49115161}{RGB}{49,115,161}

\begin{axis}[
axis line style={darkslategray38},
tick align=outside,
tick pos=left,
x grid style={lightgray204},
xlabel=\textcolor{darkslategray38}{Backbone},
xlabel style={font=\scriptsize}, 
xmin=-0.5, xmax=3.5,
xtick style={color=darkslategray38},
xtick={0,1,2,3},
xticklabel style={rotate=0.0, font=\scriptsize}, 
xticklabels={TCN,RNN FCN,mWDN,InceptionTime},
legend style={at={(0,1)}, anchor=north west, nodes={scale=0.5, transform shape}, font=\small}, 
legend cell align=left,
smooth,
y grid style={lightgray204},
ylabel=\textcolor{darkslategray38}{NCC accuracy},
ylabel style={font=\scriptsize}, 
ymajorgrids,
ymin=0.6, ymax=0.68,
ytick style={color=darkslategray38},
yticklabel style={font=\scriptsize}, 
width=0.9\textwidth, height=0.5\textwidth,
]
\draw[draw=white,fill=peru22412844,line width=0.32pt] (axis cs:0,0) rectangle (axis cs:-0.2,0.621518408744956);
\draw[draw=white,fill=peru22412844,line width=0.32pt] (axis cs:1,0) rectangle (axis cs:0.8,0.631102543388569);
\draw[draw=white,fill=peru22412844,line width=0.32pt] (axis cs:2,0) rectangle (axis cs:1.8,0.640454320102535);
\draw[draw=white,fill=peru22412844,line width=0.32pt] (axis cs:3,0) rectangle (axis cs:2.8,0.630160900322637);
\addlegendimage{ybar,ybar legend,draw=white,fill=peru22412844,line width=0.32pt}
\addlegendentry{DTAN}

\draw[draw=white,fill=steelblue49115161,line width=0.32pt] (axis cs:1,0) rectangle (axis cs:1.2,0.656342056075144);
\draw[draw=white,fill=steelblue49115161,line width=0.32pt] (axis cs:2,0) rectangle (axis cs:2.2,0.655250724125139);
\draw[draw=white,fill=steelblue49115161,line width=0.32pt] (axis cs:3,0) rectangle (axis cs:3.2,0.670476629324275);
\draw[draw=white,fill=steelblue49115161,line width=0.32pt] (axis cs:2.77555756156289e-17,0) rectangle (axis cs:0.2,0.636244756503774);
\addlegendimage{ybar,ybar legend,draw=white,fill=steelblue49115161,line width=0.32pt}
\addlegendentry{MT-DTAN}

\end{axis}
\end{tikzpicture}
\caption{DTAN vs. MT-DTAN comparison across different backbones on 113 datasets of the UCR archive~\cite{Dau:2019:ucr} (2019 version) 
evaluated by the Nearest Centroid Classification (NCC) accuracy.}
\label{fig:barplot}
\end{figure}
\begin{figure}[t!]
\centering
\def\figwidth{0.3\linewidth}

\begin{tikzpicture}
    \node[inner sep=0] (img1) at (0,0) {
        \includegraphics[trim = 9mm 9mm 12mm 15mm, clip, width=\figwidth]{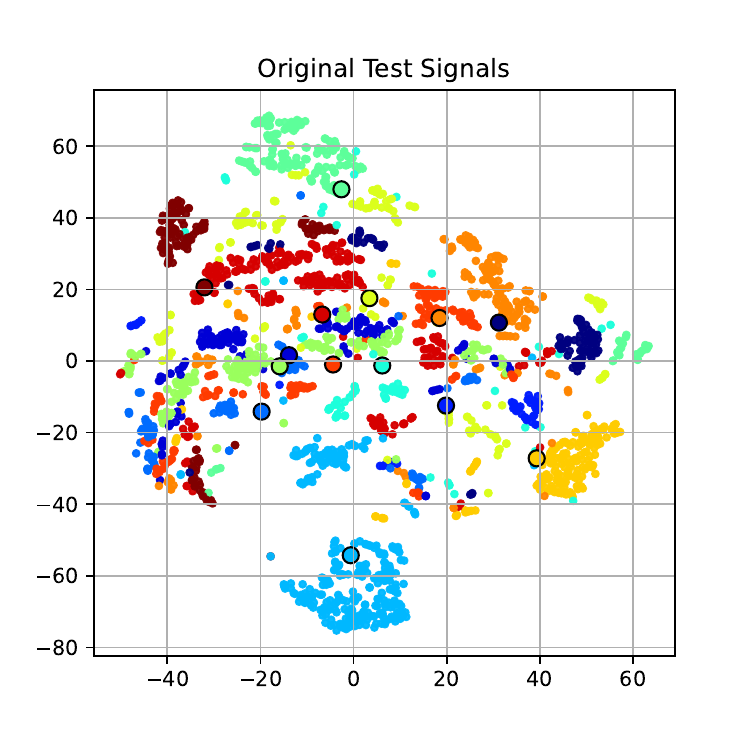}
    };
    \node[inner sep=0, right=2mm of img1] (img2) {
        \includegraphics[trim = 9mm 9mm 12mm 15mm, clip, width=\figwidth]{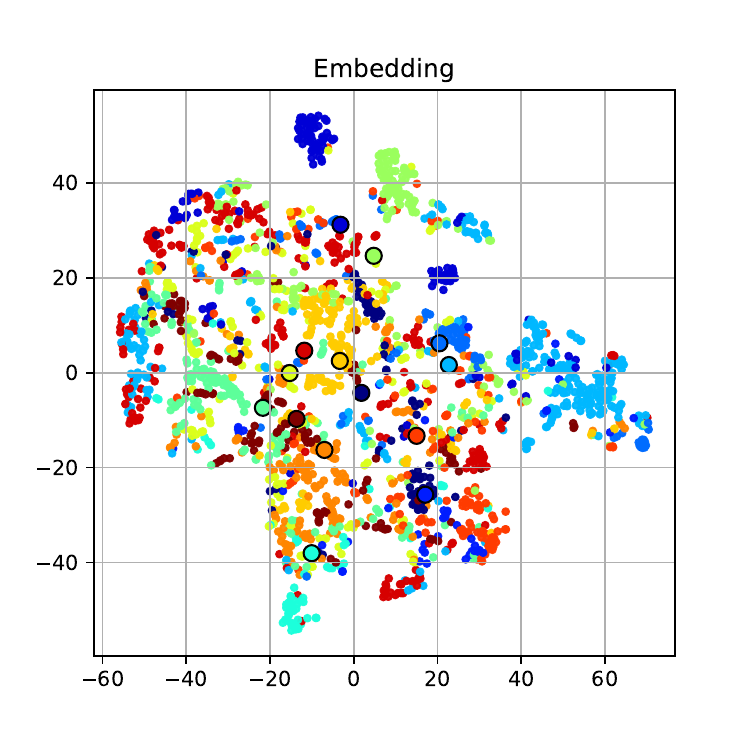}
    };
    \node[inner sep=0, right=2mm of img2] (img3) {
        \includegraphics[trim = 9mm 9mm 12mm 15mm, clip, width=\figwidth]{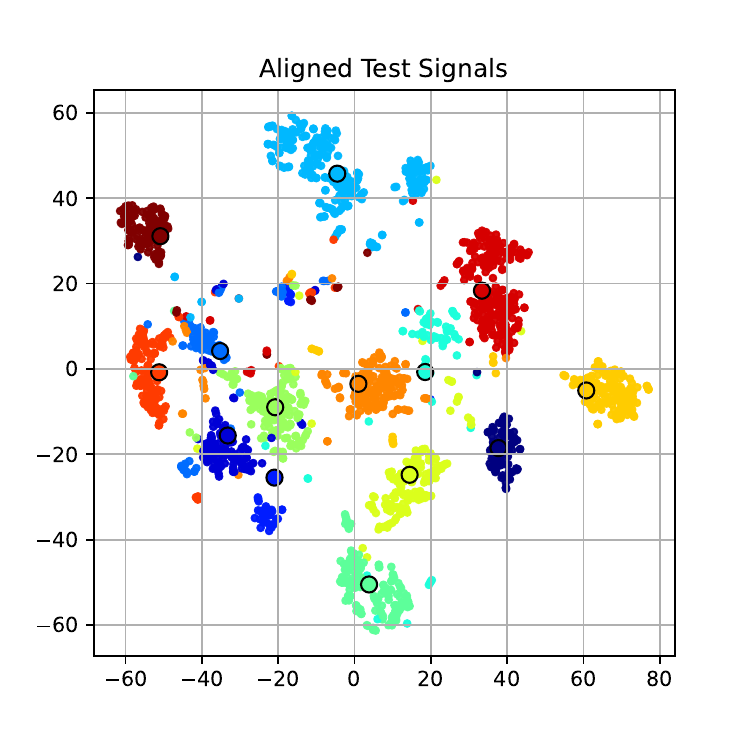}
    };

    \node[rotate=90, left=2mm of img1, anchor=center, font=\scriptsize] {DTAN};

    \node[inner sep=0, below=2mm of img1] (img4) {
        \includegraphics[trim = 9mm 9mm 12mm 15mm, clip, width=\figwidth]{figures/tsne/facesUCR/Original.pdf}
    };
    \node[inner sep=0, below=2mm of img2] (img5) {
        \includegraphics[trim = 9mm 9mm 12mm 15mm, clip, width=\figwidth]{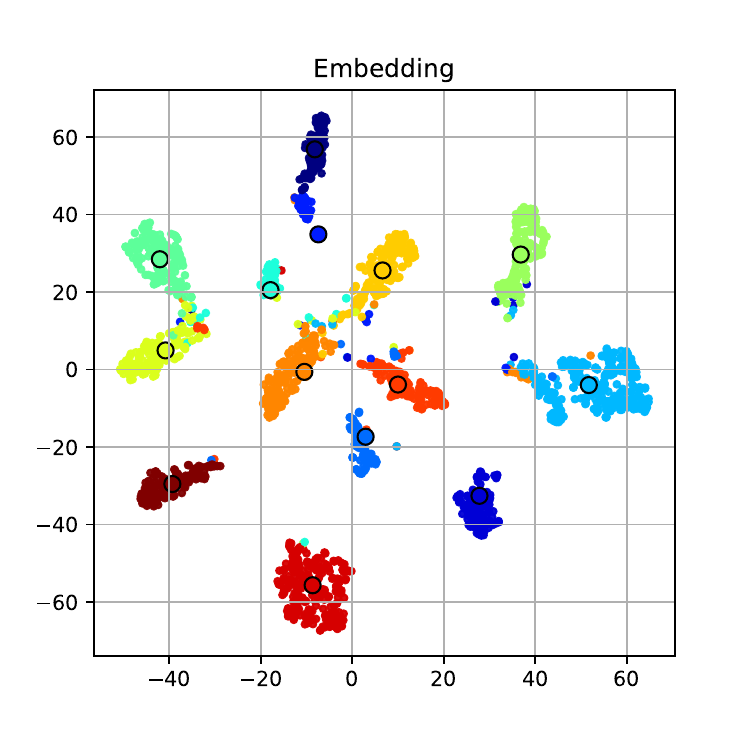}
    };
    \node[inner sep=0, below=2mm of img3] (img6) {
        \includegraphics[trim = 9mm 9mm 12mm 15mm, clip, width=\figwidth]{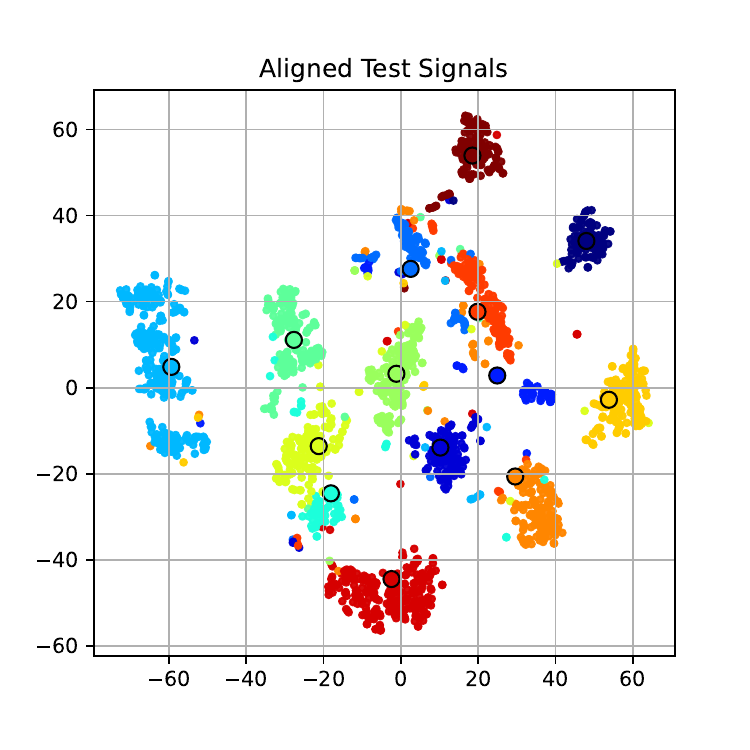}
    };

    \node[rotate=90, left=2mm of img4, anchor=center, font=\scriptsize] {MT-DTAN};
    \node[below=2mm of img4, anchor=center, font=\scriptsize] {Original data};
    \node[below=2mm of img5, anchor=center, font=\scriptsize] {Original data embedding};
    \node[below=2mm of img6, anchor=center, font=\scriptsize] {Aligned data};
\end{tikzpicture}
    \caption{t-SNE visualizations of the 14-class \textit{FacesUCR} dataset are shown for DTAN (top) and MT-DTAN (bottom) using the \emph{InceptionTime} backbone. DTAN effectively reduces the within-class variance in the original signal's domain but fails to achieve similar results for latent features, also known as the embedding. Conversely, the multi-tasking framework, MT-DTAN, demonstrates improved separation both in the original domain and in the embedding space.}
\label{fig:tsne}
\end{figure}
In this section, evaluation was performed on 113 (out of 128) datasets of the updated UCR archive~\cite{Dau:2019:ucr} (\ie, using all datasets,
omitting the ones containing VL and/or too short for the max-pooling operators in some of the architectures). Again, we used the provided train/test splits given by the authors
 of the archive and used 20\% of the train set as validation for choosing the best epoch.
The experiments in the previous sections provided an in-depth evaluation of DTAN, given a fixed $f_{\mathrm{loc}}$, across different numbers of recurrences, HP values, and objective functions. In this section, we fix
$\lambda_{\sigma}=0.1, \lambda_{\mathrm{smooth}}=0.5$, and focus on how the choice of $f_{\mathrm{loc}}$ affects DTAN's performance and the effect of multitask learning. 
To this end, we chose the following 
architectures:
\begin{enumerate}
    \item \textbf{TCN}: Temporal Convolutional Network, 
    identical to the CNN from previous sections, but with an
    adaptive average pooling operator before the penultimate layer to maintain a fixed number of parameters \wrt the input's length.
    \item \textbf{RNN-FCN:} A Recurrent Neural Network (RNN) with a hidden layer of size=100 followed by a Fully-Convolutional layer (FCN)
    with 3 blocks of [128, 256, 128] and a kernel size of [7, 5, 3] respectively.
    \item \textbf{InceptionTime}~\cite{Ismail:2020:inceptiontime}: composed of 5 Inception blocks (identical to the one used in previous sections). 
    \item \textbf{mWDN}~\cite{wang:SIGKDD:multilevel:2018}: composed of 3 Wavelet Blocks and an \emph{InceptionTime} module for the 
    \emph{Residual Alignment Flow} (RAC) framework. 
\end{enumerate}
 We have used the \emph{tsai} PyTorch implementation for RNN-FCN, mWDN, and InceptionTime~\cite{Ignacio:tsai}. 

\textbf{Results:} 
\autoref{fig:barplot} shows the average NCC test accuracy of DTAN and MT-DTAN for the different architectures on 113 datasets of the 
UCR archive~\cite{Dau:2019:ucr}.  The overall 
best performance is achieved by MT-DTAN coupled with the \emph{InceptionTime} architecture. 
This is consistent with the results presented in~\cite{Ismail:2020:inceptiontime}, where the authors show that 
\emph{InceptionTime} produced the best performance for TSC. It is therefore unsurprising that 
it presented the largest performance gain when trained in the multi-task framework, as it was specifically designed
for classification. We also note that the \emph{mWDN} architecture provides the best performance for DTAN. Overall, the results indicate the importance of the choice of $f_{\mathrm{loc}}$ when it comes to time-series JA and deep TSC architectures are a good choice for this task.

Additionally, \autoref{fig:tsne} presents the t-SNE visualization of the \textit{FacesUCR} dataset
 original data, learned embeddings, and the aligned data (using the \emph{InceptionTime} architecture).
Both DTAN and MT-DTAN alignment are sufficient for the t-SNE algorithm to provide 
adequate clustering in a 2-dimensional projection. The same cannot be said for the embeddings of the \emph{InceptionTime} model, 
as DTAN is unable to provide good separation in the latent space. In comparison, MT-DTAN can provide a good 
separation between classes for both the aligned data and its embeddings. This helps to shed light on the 
performance gains achieved by MT-DTAN compared with the original model. 

\subsection{Principal Components Analysis}~\label{Sec:Results:sub:PCA}
As discussed in~\autoref{Sec:Introduction}, time-series data pose particular issues when it comes to 
dimensionality reduction by, \eg, PCA.
While the relation between time-series data and PCA has been researched in the context of functional data analysis 
(\eg functional-PCA~\cite{dauxois:1982:asymptotic, ramsay:1991:some}) or neural computation~\cite{williams:2020:twpca}, we focus on the effect of JA
on the traditional PCA algorithm.
In particular, after JA has been learned, PCA can 
be applied to the aligned time series to produce misalignment-robust principal components (PCs).
Given a set of observations and the predicted warping parameters by DTAN, $(u_i,\btheta_i)_{i=1}^N$ respectively, 
we apply PCA on the warped data.
Given the Singular Value Decomposition (SVD) of $(v_i)_{i=1}^N$, 
\begin{align}
    (v_i)_{i=1}^N = \bP\Lambda \bQ^T= \sum_{i}^{N}\sqrt{s_i}p_{i,j}q_i\, ,
\end{align}
one can perform data reconstruction in its original domain, given the first $k$ PCs: 
\begin{align}  
    \tilde{v}_j=\sum_{i}^{k}\sqrt{s_i}p_{i,j}q_i \\
    \tilde{u}_j=\tilde{v}_j\circ {T^{-\btheta_j}}
\end{align} 

To evaluate the effect of DTAN JA on dimensionality reduction we provide results on the Trace dataset as a case study.~\autoref{fig:pca} (top) shows the cumulative explained variance by 
the first 10 PCs on both the original and aligned data. Since DTAN is set to minimize the within-class variance
by reducing temporal variability, fewer PCs are required to explain the overall variance of the entire set \wrt the original data. This is also reflected in~\autoref{fig:pca} middle-to-bottom panels, where 1) the first three PCs are presented and 2) 
the reconstruction, performed by projecting the aligned data onto the first 6 PCs. 
\textbf{The first PC of the aligned data already explains $95.7\%$ of the variance while the first three PCs of the original 
data combined explain only $88.9\%$.}
The bottom panels demonstrate that, in contrast to the misaligned original data, 6 PCs adequately reconstruct the aligned data with high fidelity and also serve as a denoising procedure, suggesting that PCA and similar dimensionality reduction methods could be enhanced by DTAN.

\section{Conclusion}\label{Sec:Conclusion}
In this work, we introduced DTAN, a novel deep learning framework for time-series Joint Alignment (JA), drawing upon a blend of contemporary and traditional concepts such as Spatial Transformer Networks (STN; ~\cite{Jaderberg:NIPS:2015:spatial,Skafte:CVPR:2018:DDTN}), efficient and highly-expressive diffeomorphisms~\cite{Freifeld:ICCV:2015:CPAB,Freifeld:PAMI:2017:CPAB}, and JA cost functions~\cite{Learned:PAMI:2006:align,Cox:CVPR:2008:LS,Erez:2022:ECCV:MCBM,Barel:ECCV:2024:spacejam}. DTAN facilitates unsupervised learning of alignments, with an extension to a weakly-supervised regime when class labels are available, enabling class-specific JA. The inherent challenges of unsupervised JA, particularly the risk of trivial solutions through excessive signal distortion, are mitigated through two distinct strategies: a regularization term for warps and the introduction of the Inverse Consistency Averaging Error (ICAE), a novel, regularization-free approach. Furthermore, we present RDTAN, an enhanced recurrent version of DTAN, which surpasses the original in terms of expressiveness and performance without an increase in parameter count. To augment class separation, we propose MT-DTAN, a multi-tasking extension of DTAN, optimized for simultaneous alignment and classification, alongside the Inverse Consistent Centroids Triplet Loss, derived from ICAE. Further evaluations demonstrate that architectures originally designed for time-series classification, when used as the backbone for the TTN, are equally effective in facilitating time-series alignment tasks. 
Finally, our findings underscore the efficacy of JA in enabling misalignbment-robust and efficient Principal Component Analysis (PCA) for time-series data.

\section*{Acknowledgements}
This work was supported by the Lynn and William Frankel Center at BGU CS, by the Israeli Council for Higher Education via the BGU Data Science Research Center, and by the Israel Science Foundation Personal Grant \#360/21. R.S.W.~was also funded in part by the BGU Kreitman School Negev Scholarship.

\ifCLASSOPTIONcaptionsoff
  \newpage
\fi

\bibliography{refs}
\bibliographystyle{IEEEtranN}
\end{document}